\documentclass[10pt,twocolumn,twoside]{IEEEtran}

\usepackage{url}
\usepackage[caption=false]{subfig}
\usepackage[american]{babel}
\usepackage{graphicx}
\usepackage{amsmath,amssymb} 
\usepackage{algorithm,algorithmic}
\usepackage{changepage}

\usepackage[pagebackref=true,breaklinks=true,colorlinks,bookmarks=false]{hyperref}

\begin{document}
\title{A Diffusion and Clustering-based Approach for Finding Coherent Motions and Understanding Crowd Scenes}

\author{Weiyao Lin,~Yang Mi,~Weiyue Wang,~Jianxin Wu,~Jingdong Wang,~and Tao Mei%
\thanks{This paper is supported in part by the following grants: Chinese National 973 Grants (2013CB329603), National Science Foundation of China (No. 61471235, 61422203, 61425011, 61527804), and Microsoft Research Asia Collaborative Research Award. The basic idea of this paper appeared in our
conference version \cite{40}. In this version, we propose a new cluster-and-merge process to mine recurrent activities in crowd scenes, carry out detailed analysis, and present more performance results.}
\thanks{W. Lin, Y. Mi and W. Wang are with the Department of Electronic Engineering, Shanghai Jiao Tong University, China (email: \{wylin, deyangmiyang, laughter\}@sjtu.edu.cn). }
\thanks{J. Wu is with the National Key Laboratory for Novel Software Technology, Nanjing University, China (email: wujx2001@nju.edu.cn).}
\thanks{J. Wang and T. Mei are with the Microsoft Research, Beijing, China (email: \{jingdw, tmei\}@microsoft.com).}
}


\maketitle

\begin{abstract}
This paper addresses the problem of detecting coherent motions in crowd scenes and presents its two applications in crowd scene understanding: semantic region detection and recurrent activity mining. It processes input motion fields (e.g., optical flow fields) and produces a coherent motion filed, named as thermal energy field. The thermal energy field is able to capture both motion correlation among particles and the motion trends of individual particles which are helpful to discover coherency among them. We further introduce a two-step clustering process to construct stable semantic regions from the extracted time-varying coherent motions. These semantic regions can be used to recognize pre-defined activities in crowd scenes. Finally, we introduce a cluster-and-merge process which automatically discovers recurrent activities in crowd scenes by clustering and merging the extracted coherent motions. Experiments on various videos demonstrate the effectiveness of our approach.
\end{abstract}


\section{Introduction}

\begin{figure}[t]
  \centering
  \subfloat[]{\includegraphics[height=2.0cm]{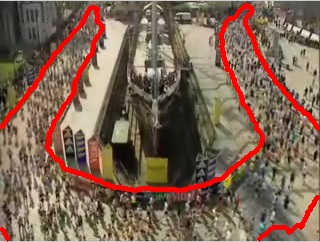}\label{fig:coherent_example_a}}
  \hspace{2mm}
  \subfloat[]{\includegraphics[width=0.15\textwidth]{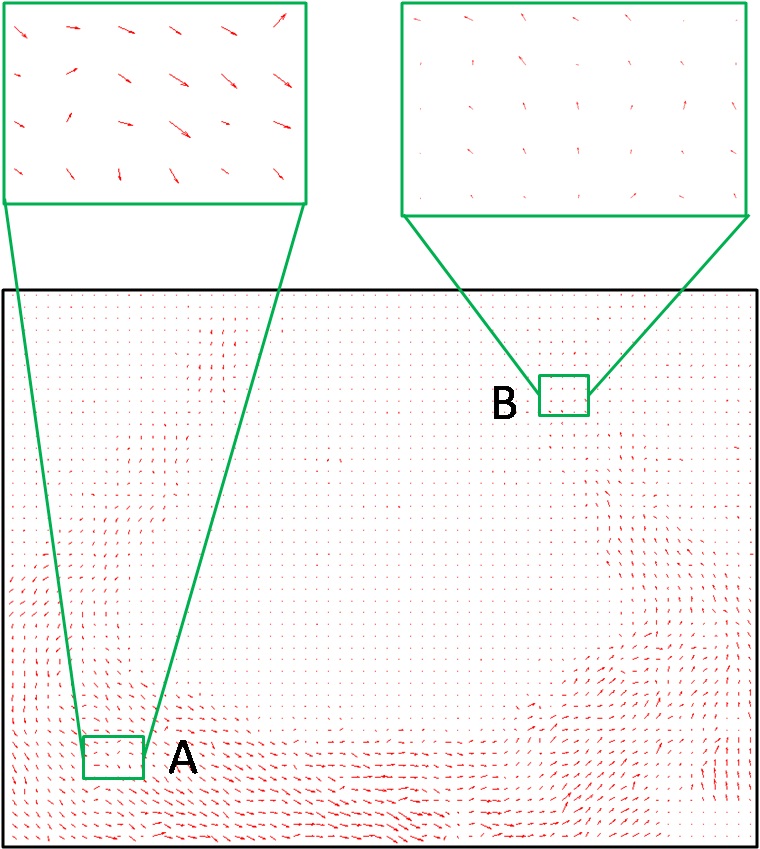}\label{fig:coherent_example_b}}
  \hspace{2mm} \subfloat[]{\includegraphics[width=0.15\textwidth]{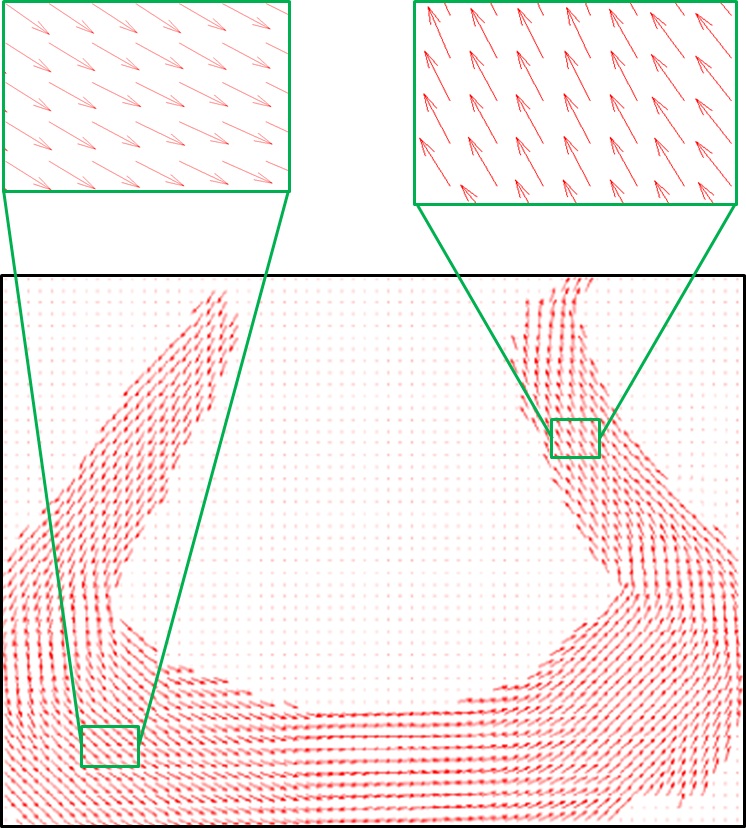}\label{fig:coherent_example_c}}
  \caption{(a) Example frame of a Marathon video sequence, the red curve is the coherent motion region; (b) Input motion vector field of (a); (c) Coherent motion field from (b) using the proposed approach (Best viewed in color).}\label{fig:coherent_example}
\end{figure}

\begin{figure}[t]
  \centering
  \subfloat[]{\includegraphics[width=2.6cm,height=2.6cm]{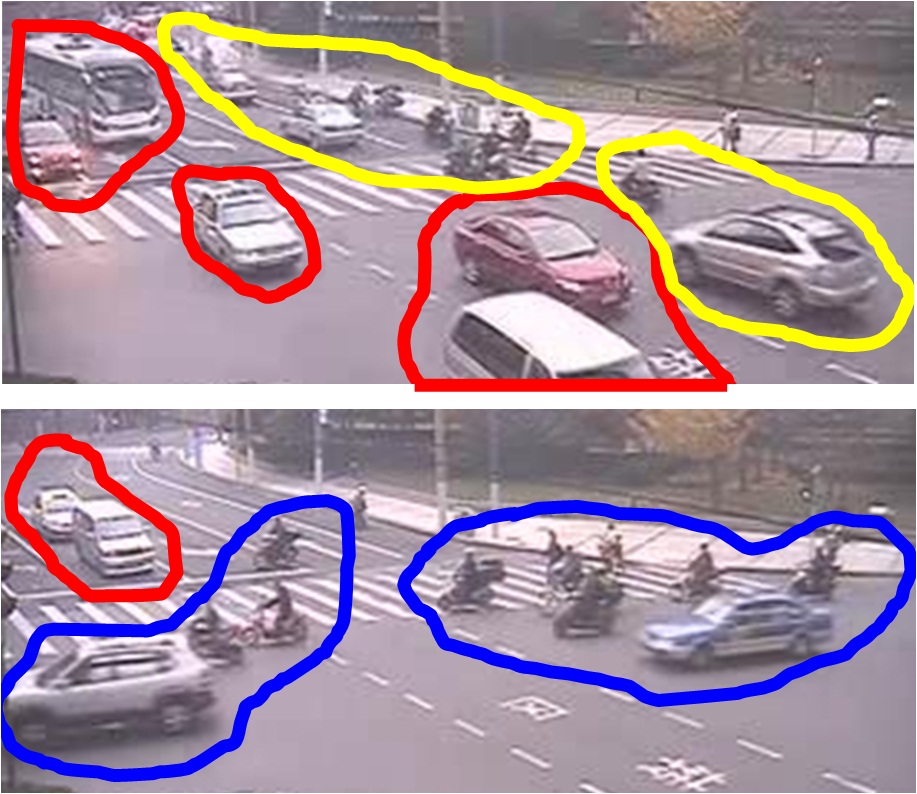}\label{fig:process_example_a}}
   \hspace{2mm}
  \subfloat[]{\includegraphics[width=2.6cm,height=1.25cm]{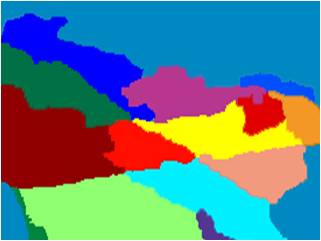}\label{fig:process_example_b}}
   \hspace{2mm}
  \subfloat[]{\includegraphics[width=2.6cm,height=2.6cm]{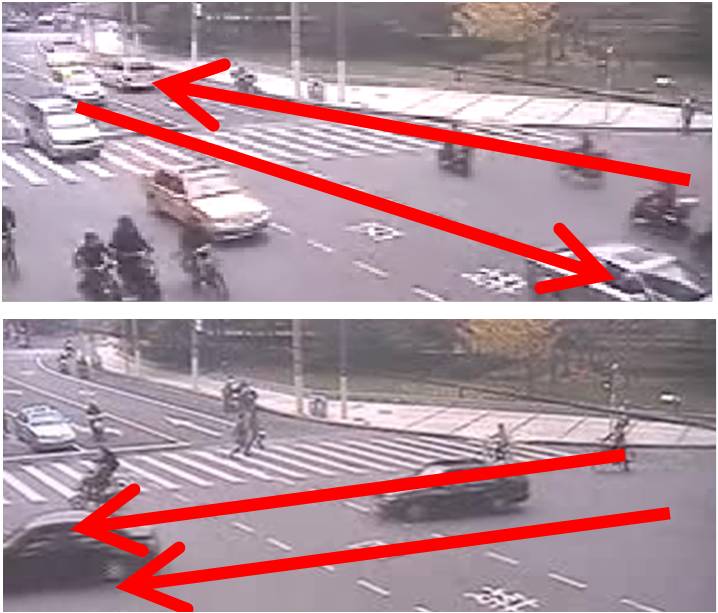}\label{fig:process_example_c}}
  \caption{(a) Example time-varying coherent motions in a scene, where different coherent motions are circled by curves with different color; (b) Constructed semantic regions for the scene in (a); (c) Recurrent activities for the scene in (a), where the arrows represent the major motion flows in each recurrent activity (Best viewed in color).}\label{fig:process_example}
\end{figure}

Coherent motions, which represent coherent movements of massive individual particles, are pervasive in natural and social scenarios. Examples include traffic flows and parades of people (cf. Figs~\ref{fig:coherent_example_a} and~\ref{fig:process_example_a}). Since coherent motions can effectively decompose scenes into meaningful semantic parts and facilitate the analysis of complex crowd scenes, they are of increasing importance in crowd-scene understanding and activity recognition \cite{8,20,22,37,41}.

In this paper, we address the problem of detecting coherent motions in crowd scenes, and subsequently using them to understand input scenes. More specifically, we focus on 1) constructing an accurate coherent motion field to find coherent motions, 2) finding stable semantic regions based on the detected coherent motions and using them to recognize pre-defined activities (i.e., activities with labeled training data) in a crowd scene, and 3) automatically mining recurrent activities in a crowd scene based on the detected coherent motions and semantic regions.

First, constructing an accurate coherent motion field is crucial in detecting reliable coherent motions. In Fig.~\ref{fig:coherent_example},~\subref{fig:coherent_example_b} is the input motion field and~\subref{fig:coherent_example_c} is the coherent motion field which is constructed from~\subref{fig:coherent_example_b} using the proposed approach. In~\subref{fig:coherent_example_b}, the motion vectors of particles at the beginning of the Marathon queue are far different from those at the end, and there are many inaccurate optical flow vectors. Due to such variations and input errors, it is difficult to achieve satisfying coherent motion detection results directly from~\subref{fig:coherent_example_b}. However, by transferring~\subref{fig:coherent_example_b} into a coherent motion field where the coherent motions among particles are suitably highlighted in~\subref{fig:coherent_example_c}, coherent motion detection is greatly facilitated. Although many algorithms have been proposed for coherent motion detection \cite{2,3,7,8}, this problem is not yet effectively addressed. \emph{We argue that a good coherent motion field should effectively be able to} 1) \emph{encode motion correlation among particles}, such that particles with high correlations can be grouped into the same coherent region; and, 2) \emph{maintain motion information of individual particles}, such that activities in crowd scenes can be effectively parsed by the extracted coherent motion field. Based on these intuitions, we propose a thermal-diffusion-based approach, which can extract accurate coherent motion fields.

Second, constructing meaningful semantic regions to describe activity patterns in a scene is also essential. Coherent motions at different times may vary widely. In Fig.~\ref{fig:process_example_a}, changing of traffic lights will lead to different coherent motions. Coherent motions alone may not effectively describe the overall semantic patterns in a scene either. Therefore, semantic regions need to be extracted from these time-varying coherent motions to achieve stable and meaningful semantic patterns, as in Fig.~\ref{fig:process_example_b}. However, most existing works only focus on the detection of coherent motions at some specific time, while the problem of handling time-varying coherent motions is less studied. We proposed a two-step clustering process for this purpose.

Third, mining recurrent activities is another important issue. Many crowd scenes are composed of recurrent activities \cite{31,32,33}. For example, the scene in Fig.~\ref{fig:process_example} is composed of recurrent activities including vertical motion activities and horizontal motion activities, as in Fig.~\ref{fig:process_example_c}. Automatically mining these recurrent activities is important in understanding scene contents and their dynamics. Although many researches have been done for parsing recurrent activities in low-crowd scenes \cite{28,29,30,38}, this issue is not well addressed in crowd scene scenarios where reliable motion trajectories are unavailable. We proposed a cluster-and-merge process, which can effectively extract recurrent activities in crowd scenes.

Our contributions to crowd scene understanding and activity recognition are summarized as follows.
\begin{enumerate}
 \item We introduce a coarse-to-fine thermal diffusion process to transfer an input motion field into a thermal energy field (TEF), which is a more accurate coherent motion field. TEF effectively encodes both motion correlation among particles and motion trends of individual particles. To our knowledge, this is the first work that introduces thermal diffusion to detect coherent motions in crowd scenes. We also introduce a triangulation-based scheme to effectively identify coherent motion components from the TEF.
 \item We present a two-step clustering scheme to find semantic regions according to the correlations among coherent motions. The found semantic regions can effectively catch activity patterns in a scene. Thus good performance can be achieved when recognizing pre-defined crowd activities based on these semantic regions.
 \item We propose a cluster-and-merge process to automatically mine recurrent activities by clustering and merging the coherent motions. The obtained recurrent activities can accurately describe recurrent motion patterns in a crowd scene.
\end{enumerate}

The remainder of this paper is organized as follows. Section~\ref{section:related_work} reviews related works. Section~\ref{section:overview} describes the framework of the proposed approach. Sections~\ref{section:coarse_to_fine} to~\ref{section:cluster_and_merge} describe the details of our proposed thermal diffusion process, triangulation scheme, two-step clustering scheme, and cluster-and-merge process. Section~\ref{section:experiments} shows the experimental results and Section~\ref{section:conclusion} concludes the paper.

\section{Related Works\label{section:related_work}}

Many works \cite{1,2,3,7,8,4,5,10,11,9} have been proposed on coherent motion detection. Due to the complex nature of crowd scenes, they are not yet mature for accurate detection of coherent motion fields. Cremers and Soatto \cite{10} and Brox et al. \cite{11} model the intensity variation of optical flow by an objective functional minimization scheme. These methods are only suitable for motions with simple patterns and cannot effectively analyze complex crowd patterns such as the circular flow in Fig.~\ref{fig:coherent_example_a}. Other works introduce external spatial-temporal correlation traits to model the motion coherency among particles \cite{2,3,7}. Since these methods model particle correlations in more precise ways, they can achieve more satisfying results. However, most of these methods only consider short-distance particle motion correlation within a local region while neglecting long-distance correlation among distant particles, they have limitations in handling low-density or disconnected coherent motions where the long-distance correlation is essential. Furthermore, without the information from distant particles, these methods are also less effective in identifying coherent motion regions in the case when local coherent motion patterns are close to their neighboring backgrounds. One example of this kind of scenario is showcased in the region B in Fig.~\ref{fig:coherent_example_b}.

There are also other works related to motion modeling. One line of related works is advanced optical flow estimation. These methods try to improve the estimation accuracy of the input motion field by including global constraints over particles \cite{13,14,18,39}. The focus of our approach is different from these methods. We focus on enhancing the correlation among coherent particles to facilitate coherent motion detection. Thus, the motion vectors of coherent particles are enhanced even if their actual motions are small, such as the region B in Figs~\ref{fig:coherent_example_b} and~\ref{fig:coherent_example_c}. In contrast, advanced optical flow estimation methods focus on estimating the \emph{actual} motion of particles. They are still less capable of creating precise results when applied to coherent motion detection.

The anisotropic diffusion based methods, used in image segmentation, is also related to our work \cite{15,26,27}. Our approach differs from these methods. First, our approach not only embeds the motion correlation among particles, but also suitably maintains the original motion information from the input motion vector field. Comparatively, the anisotropic-diffusion-based methods are more focused on enhancing the correlation among particles while neglecting the particles' original information. As aforementioned, maintaining particle motion information is important in parsing crowd scenes. More importantly, due to the complex nature of crowd scenes, many coherent region boundaries are vague, subtle and unrecognizable. Simply applying the anisotropic-diffusion methods cannot identify the ideal boundaries. The proposed thermal diffusion process can achieve more satisfying results by modeling the motion direction, strength, and spatial correlation among particles.

Besides coherent motion detection, it is also important to utilize coherent motions to recognize pre-defined crowd activities. However, most existing coherent motion works only focus on the extraction of coherent motions while the recognition of crowd activities is much less studied. In \cite{1}, Ali and Shah detected instability regions in a scene by comparing with its normal coherent motions. However, they assume coherent motions to be stable, while in practice, many coherent motions may vary widely over time, making it difficult to construct stable normal coherent motions. Furthermore, besides the works on coherent motion, there are also other works which directly extract global features from the entire scene to recognize crowd activities \cite{20,21}. However, since they do not consider the semantic region correlations inside the scene, they have limitations in differentiating subtle differences among activities. Although there are some works \cite{22,23} which recognize crowd activities by segmenting scenes into semantic regions, our approach differs from them. Our approach finds the semantic regions by first extracting global coherent motion information, while these methods construct semantic regions from the particles' local features. As will be shown later, information from the coherent motions can effectively enhance the correlation among particles, resulting in more meaningful semantic regions to facilitate activity recognition.

Furthermore, pre-defining or labeling crowd activities requires lots of human labors, making it desirable to automatically discover activity patterns in a crowd video without human intervention. In \cite{30}, Morris and Trivedi clustered trajectories into groups and modeled the spatio-temporal dynamic patterns of each trajectory group by Hidden Markov Models. Wang et al. \cite{28} and Hu et al. \cite{29} further introduced Dirichlet processes to model the activity patterns of different trajectory groups. However, since these methods extract recurrent activities from motion trajectories, they are not suitable for crowd scene scenarios where reliable trajectories are difficult to achieve. Besides using motion trajectories, other researches tried to find recurrent activities by extracting low-level or short-term motion features. For example, Zhou et al. \cite{33} extracted fragments of trajectories (called tracklets) and utilized a Latent Dirichlet Allocation topic model to infer recurrent activities. Jagannadan et al. \cite{32} and Emonet et al. \cite{31} extracted low-level motion flows as motion descriptors and introduced a Probabilistic Latent Sequential Motif (PLSM) model to achieve recurrent activities. Although these methods can be applied in crowd scenes, they still have limitations in obtaining precise recurrent activity patterns under scenes with complex motions. Our approach differs from the previous methods in that 1) Our approach utilizes coherent motions to discover recurrent activities. Since coherent motions can effectively catch the local activity pattern in each frame, more precise recurrent activities can be achieved by our approach, 2) Our approach also extracts flow curves to describe and visualize recurrent activities. Compared with the previous methods which described recurrent activities by trajectory clusters or probability densities, the flow curves derived by our approach can visualize recurrent activity patterns in a clearer and more straightforward way.

\section{Overview of the Approach\label{section:overview}}

\begin{figure*}
\centering
  \includegraphics[width=1.0\textwidth,height=0.2\textheight]{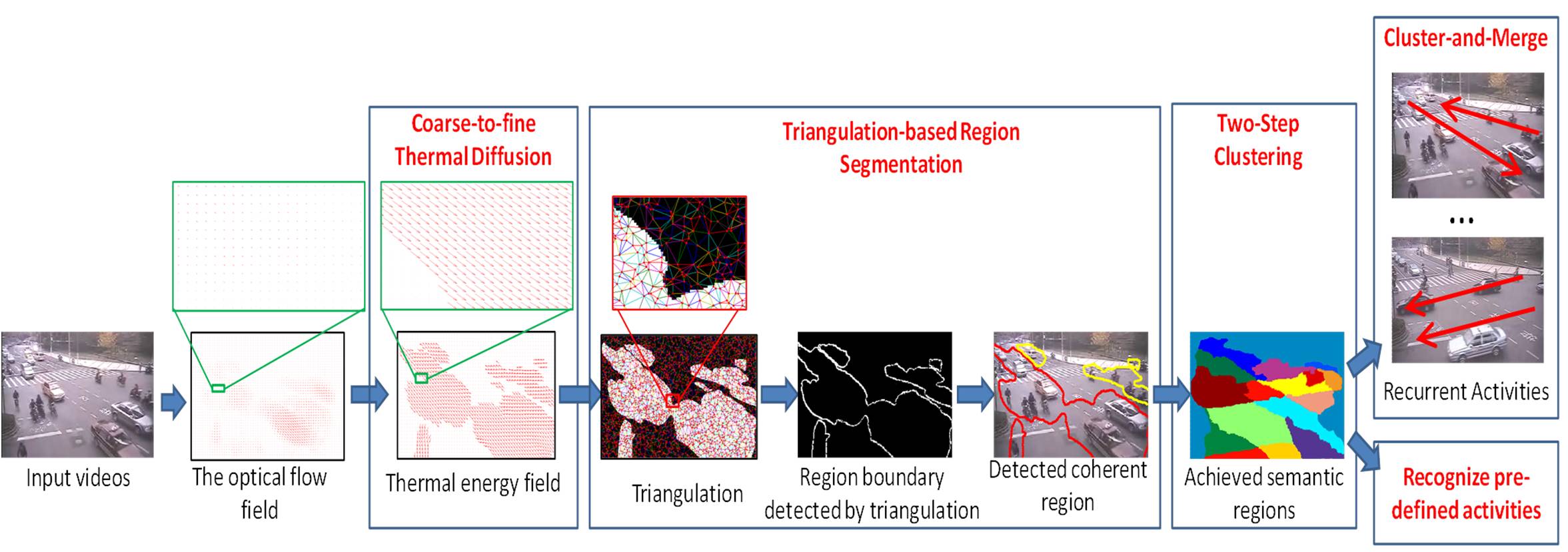}
  \caption{The flowchart of the proposed approach (best viewed in color).}\label{fig:framework}
\end{figure*}

The framework of the proposed approach is shown in Fig.~\ref{fig:framework}. The optical flow fields \cite{1,6} are first extracted from input videos. Secondly, the coarse-to-fine thermal diffusion process is applied to transfer the input motion fields into coherent motion fields, i.e., thermal energy fields (TEFs). Thirdly, the triangulation-based scheme is applied to identify coherent motions. Fourthly, with the obtained coherent motions, the two-step clustering scheme is performed to cluster coherent motions from multiple TEFs and construct semantic regions for the target scene. Finally, based on these semantic regions, we can extract effective features to describe crowd activities in the scene and recognize pre-defined crowd activities accordingly. At the same time, the cluster-and-merge process is also applied based on the extracted coherent motions and semantic regions to discover recurrent activities in the target scene. These proposed techniques are described in the following sections in detail.

\section{Finding Coherent Motions\label{section:coarse_to_fine}}

In order to find accurate coherent motions, it is important to construct a coherent motion field to highlight the motion correlation among particles while still maintaining the original motion information. To achieve this requirement, we introduce a \emph{thermal diffusion process} to model particle correlations. Given an input optical flow field, we view each particle (i.e., each pixel in a frame) as a ``heat source'' and it can diffuse energies to influence other particles. By suitably modeling this thermal diffusion process, precise correlation among particles can be achieved. The formulation is motivated by the following intuitions:
\begin{enumerate}
  \item Particles farther from heat source should achieve fewer thermal energies;
  \item Particles residing in the motion direction of the heat source particle should receive more thermal energies;
  \item Heat source particles with larger motions should carry more thermal energies.
\end{enumerate}

\subsection{Thermal Diffusion Process}

Based on the above discussions, we borrow the idea from physical thermal propagation \cite{17} and model the thermal diffusion process by Eq.~\ref{equation:eq1}:
\begin{equation}
\dfrac {\partial \mathbf{E}_{\mathbf{P},l}} {\partial l}=k_{p}^{2}\left( \dfrac {\partial ^{2}\mathbf{E}_{\mathbf{P},l}} {\partial x^{2}}+\dfrac {\partial^{2}\mathbf{E}_{\mathbf{P},l}} {\partial y^{2}}\right) +\mathbf{F}_{\mathbf{P}}\label{equation:eq1}
\end{equation}
where $\mathbf{E}_{\mathbf{P}, l}=[E^x_{\mathbf{P}, l},  E^y_{\mathbf{P}, l}]$ is the thermal energy for the particle at location $\mathbf{P}=(p^x, p^y)$ after performing thermal diffusion for $l$ seconds, $\mathbf{F}_{\mathbf{P}}=[f_{\mathbf{P}}^x, f_{\mathbf{P}}^y]$ is the input motion vector for particle $\mathbf{P}$, $k_p$ is the propagation coefficient.

The first term in Eq.~\ref{equation:eq1} models the propagation of thermal energies over free space such that the spatial correlation among particles can be properly enhanced during thermal diffusion. The second term $\mathbf{F}_{\mathbf{P}}$ can be viewed as the external force added on the particle to affect its diffusion behavior, which preserves the original motion patterns. The inclusion of this term is one of the major differences between the proposed approach and the anisotropic-diffusion methods \cite{27}. Without the $\mathbf{F}_{\mathbf{P}}$ term, Eq.~\ref{equation:eq1} can be solved by:
\begin{equation}
    \mathbf{E}_{\mathbf{P},l} =\frac{1}{wh} \sum _{\mathbf{Q}\in{\mathbf{I}}, \mathbf{Q}\neq{\mathbf{P}}}\mathbf{e}_{\mathbf{P},l}\left(\mathbf{Q}\right)\label{equation:eq2}
\end{equation}
where $\mathbf{E}_{\mathbf{P},l}$ is the final diffused thermal energy for particle $\mathbf{P}$ after $l$ seconds, $\mathbf{I}$ is the set of all particles in the frame,  $w$ and $h$ are width and height of the frame. The individual thermal energy {\small $\mathbf{e}_{\mathbf{P},l}\left(\mathbf{Q}\right)=[e^x_{\mathbf{P},l}\left(\mathbf{Q}\right), e^y_{\mathbf{P},l}\left(\mathbf{Q}\right)]$} is diffused from the heat source particle $\mathbf{Q}=(q^x, q^y)$ to particle $\mathbf{P}$ after $l$ seconds, defined as:
\begin{equation}
 e^{\gamma}_{\mathbf{P},l}\left(\mathbf{Q}\right)=u^{\gamma}_{\mathbf{Q}}\cdot e^{\frac{-k_{p}}{l}||\mathbf{P}-\mathbf{Q}||^2}\label{equation:eq3}
\end{equation}
where $\gamma\in\{x,y\}$, $\mathbf{U}_{\mathbf{Q}}=(u^x_{\mathbf{Q}}, u^y_{\mathbf{Q}})$ is the current motion pattern for the heat source particle $\mathbf{Q}$ and it is initialized by $\mathbf{U}_{\mathbf{Q}}=\mathbf{F}_{\mathbf{Q}}$, $||\mathbf{P}-\mathbf{Q}||$ is the distance between particles $\mathbf{P}$ and $\mathbf{Q}$. In this paper, we fix $l$ to be 1 to eliminate its effect.

However, when $\mathbf{F}$ in Eq.~\ref{equation:eq1} is non-zero, it is difficult to get the exact solution for Eq.~\ref{equation:eq1}. So we introduce an additional term $e^{-k_{f}|\mathbf{F}_{\mathbf{Q}}\cdot\left(\mathbf{P}-\mathbf{Q}\right)|}$ to approximate the influence of $\mathbf{F}_{\mathbf{Q}}$ where $k_f$ is a force propagation factor. Moreover, in order to prevent unrelated particles from accepting too much heat from $\mathbf{Q}$, we restrict that only highly correlated particles will propagate energies to each other. The final individual thermal energy from $\mathbf{Q}$ to $\mathbf{P}$ is:
\begin{equation}
 e^{\gamma}_{\mathbf{P},l}\left(\mathbf{Q}\right) = u^{\gamma}_{\mathbf{Q}}\times e^{-k_{p}||\mathbf{P}-\mathbf{Q}||^2}\times e^{-k_{f}|\mathbf{F}_{\mathbf{Q}}\cdot\left(\mathbf{P}-\mathbf{Q}\right)|}\label{equation:eq4}
\end{equation}
if $\cos(\mathbf{F}_{\mathbf{P}},\mathbf{F}_{\mathbf{Q}})\geq \theta_{c}$ and is 0 if otherwise, where $\mathbf{F}_{\mathbf{P}}$ and $\mathbf{F}_{\mathbf{Q}}$ are the input motion vectors of the current particle $\mathbf{P}$ and the heat source particle $\mathbf{Q}$, and $\cos(\mathbf{F}_{\mathbf{P}},\mathbf{F}_{\mathbf{Q}})$ is the cosine similarity, $\theta_{c}$ is a threshold. In our experiments, $k_p$, $k_f$, and $\theta_{c}$ are set to be 0.2, 0.8, 0.7, which are decided from the experimental statistics.

From Eq.~\ref{equation:eq2}, we see that the diffused thermal energy $\mathbf{E}_{\mathbf{P}}$ is the summation from all other particles, which encodes the correlation among $\mathbf{P}$ and all other particles in the frame. Furthermore, in Eq.~\ref{equation:eq4}, the first term preserves the motion pattern of the heat source. The second term considers the spatial correlation between source and target particles. The third term guarantees that particles along the motion direction of the heat source receives more thermal energies. Furthermore, the cosine similarity $\cos(\mathbf{F}_{\mathbf{P}},\mathbf{F}_{\mathbf{Q}})$ is introduced in Eq.~\ref{equation:eq4} such that particle $\mathbf{P}$ will not accept energy from $\mathbf{Q}$ if their input motion vectors are far different (or less-coherent) from each other. That is, Eq.~\ref{equation:eq4} successfully satisfies all the intuitions.

\begin{figure}
 \centering
 \subfloat[]{\includegraphics[width=0.23\linewidth]{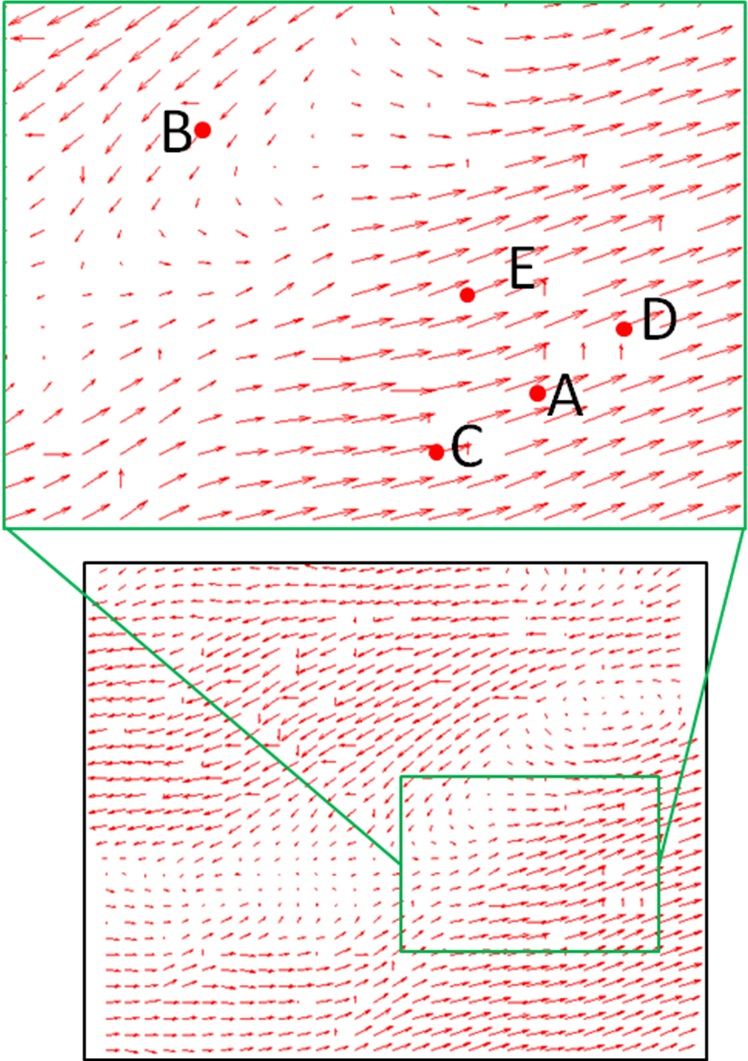}\label{fig:TEF_example_a}}
 \hspace{2mm} \subfloat[]{\includegraphics[width=0.23\linewidth]{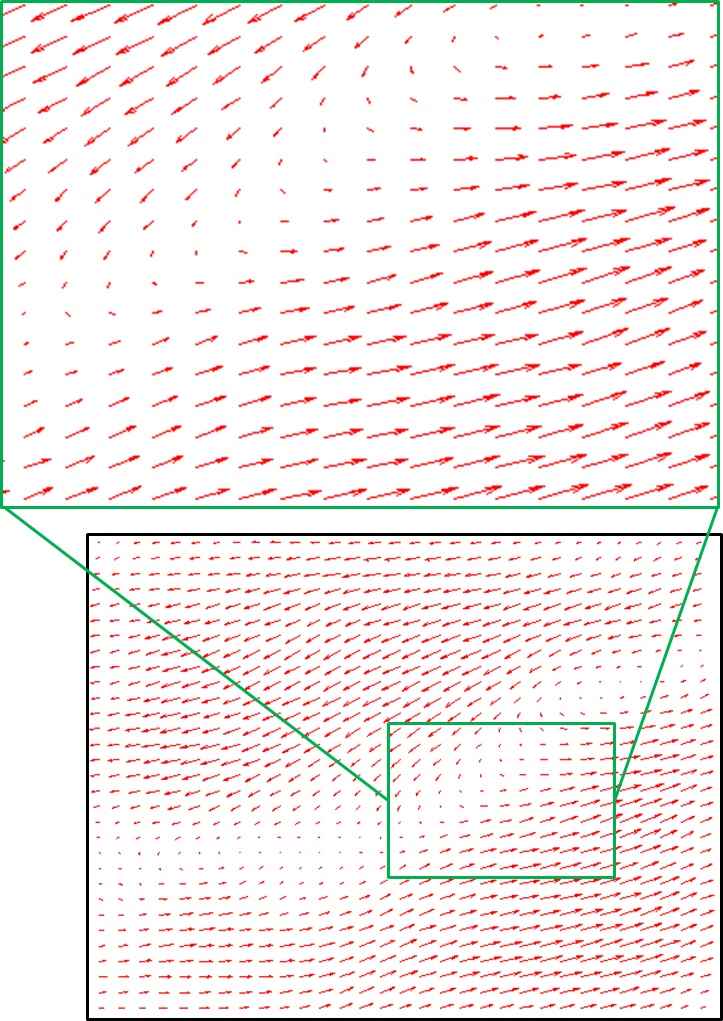}\label{fig:TEF_example_b}}
 \hspace{2mm} \subfloat[]{\includegraphics[width=0.215\linewidth]{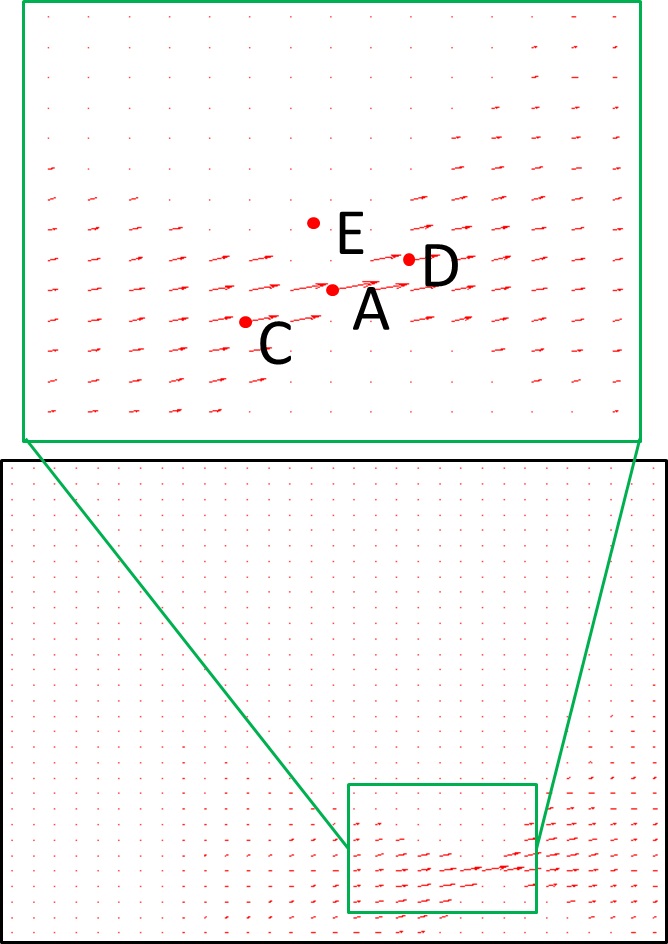}\label{fig:TEF_example_c}}
 \hspace{2mm} \subfloat[]{\includegraphics[width=0.215\linewidth]{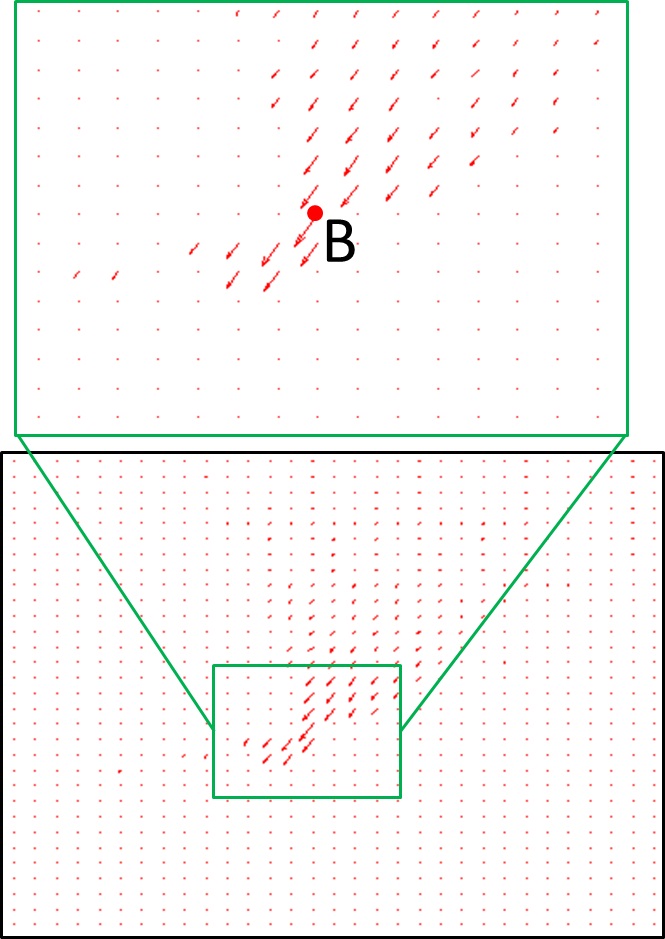}\label{fig:TEF_example_d}}
 \caption{(a),(b): One input optical flow field and its thermal energy field; (c), (d): Individual thermal diffusion result by diffusing from a single heat source particle A and B to the entire field.}\label{fig:TEF_example}
\end{figure}

Fig.~\ref{fig:TEF_example} shows one example of the thermal diffusion process, which reveals that:
\begin{enumerate}
 \item Comparing Figs~\ref{fig:TEF_example_b} and~\ref{fig:TEF_example_a}, the original motion information is indeed preseved in the TEF. Moreover, TEF further strengthens particle motion coherency by thermal diffusion, which integrates the influence among particles. Coherent motions become more recognizable, thus more accurate coherent motion extraction can be achieved.
 \item From Fig.~\ref{fig:TEF_example_c}, we can see that the thermal energy for each heat source particle is propagated in a sector shape. Particles along the motion direction of the heat source (C and D) receive more energies than particles outside the motion direction (such as E). In Fig.~\ref{fig:TEF_example_d}, since particles on the lower side of the heat source B have small (cosine) motion similarities with B, they do not accept thermal energies.
\end{enumerate}

\subsection{The Coarse-to-Fine Scheme}

Although Eqs~\ref{equation:eq2} and~\ref{equation:eq4} can effectively strengthen the coherency among particles, it is based on a single input motion field, and only short-term motion information is considered, which is volatile and noisy. Thus, we propose a coarse-to-fine scheme to include long-term motion information. The entire coarse-to-fine thermal diffusion process is described in Algorithm~\ref{algorithm:coarse_to_fine}.

\begin{algorithm}
 \caption{Coarse-to-Fine Thermal Diffusion Process}
 \begin{algorithmic}[1]
 \small{
  \STATE $T=T_{max}$
  \STATE Calculate the input motion vector field $\mathbf{F}_{\mathbf{P}}(T)$ with $T$-frame intervals
  \STATE $\mathbf{U}_{\mathbf{P}}= \mathbf{F}_{\mathbf{P}}(T)$
  \FOR{$n=0$ to $Num_{itr}$}
    \STATE Use Eq.~\ref{equation:eq2} to create the new thermal energy field $\mathbf{E}^{n}_{\mathbf{P}}$ based on $\mathbf{F}_{\mathbf{P}}(T)$ and $\mathbf{U}_{\mathbf{P}}$
    \STATE Normalize the vector magnitudes in $\mathbf{E}^{n}_{\mathbf{P}}$
    \STATE $\mathbf{U}_{\mathbf{P}}= \mathbf{E}^{n}_{\mathbf{P}}$
    \STATE $T=T-T_{step}$
    \IF{$T>0$}
      \STATE Calculate $\mathbf{F}_{\mathbf{P}}(T)$ with the new $T$
    \ENDIF
  \ENDFOR
  \STATE Output $\mathbf{E}^{n}_{\mathbf{P}}$}
 \end{algorithmic}\label{algorithm:coarse_to_fine}
\end{algorithm}

\begin{figure}
 \centering
 \subfloat[]{\includegraphics[width=2.4cm]{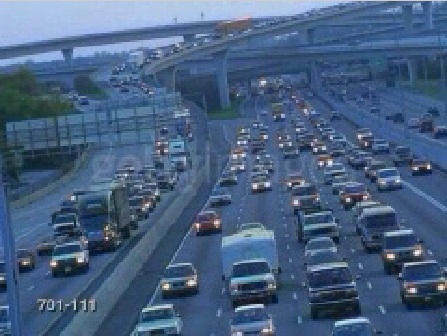}\label{fig:TEF_result_a}}
 \hspace{6pt}
 \subfloat[]{\includegraphics[width=2.4cm]{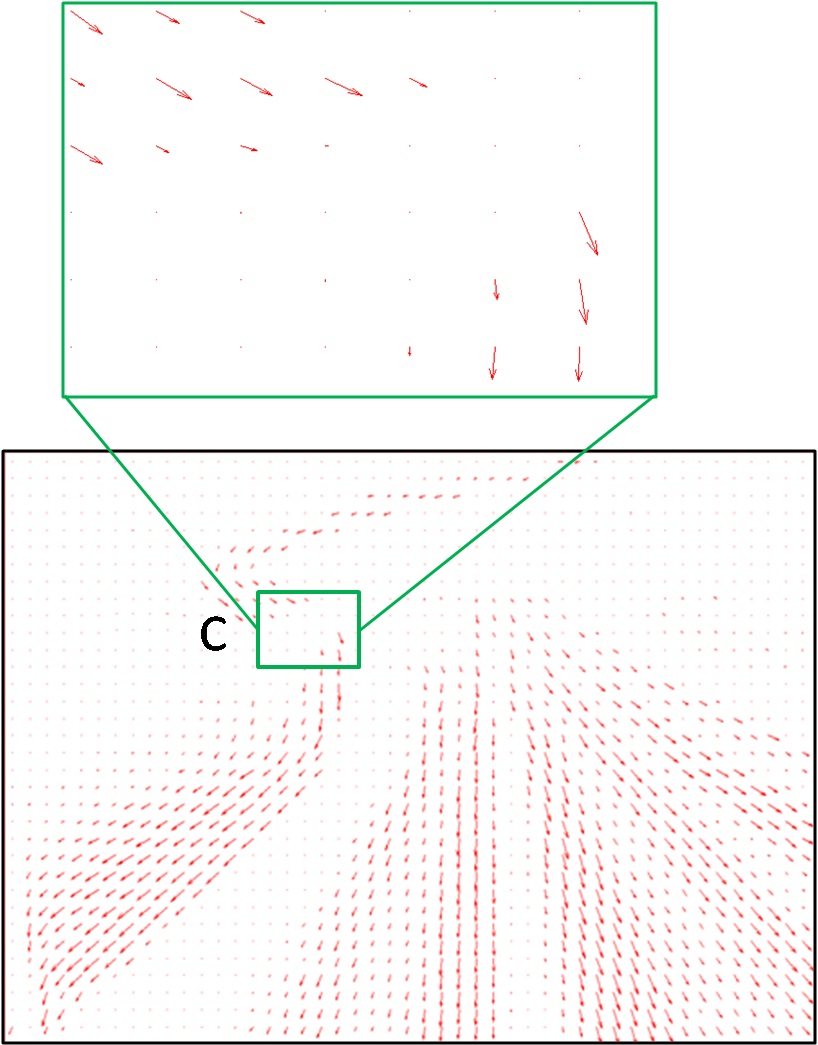}\label{fig:TEF_result_b}}  \\
 \subfloat[]{\includegraphics[width=2.4cm]{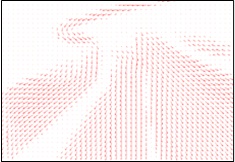}\label{fig:TEF_result_c}}
 \hspace{6pt}
 \subfloat[]{\includegraphics[width=2.4cm]{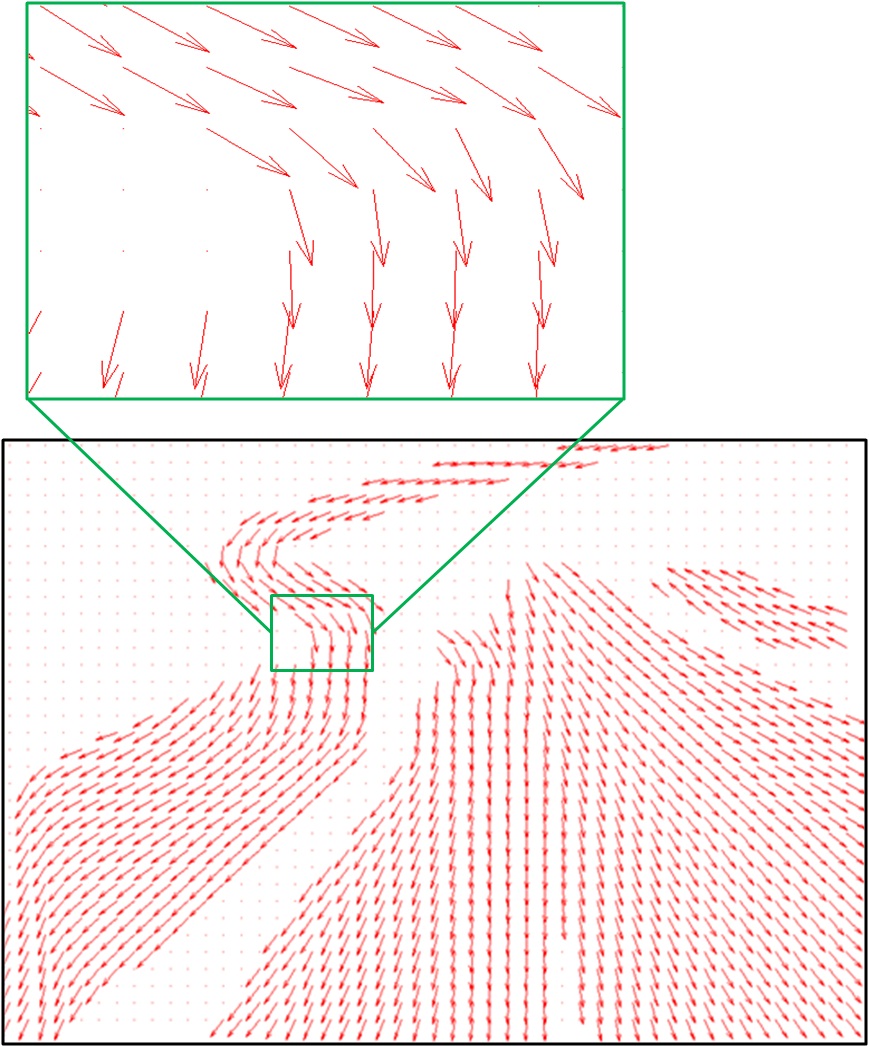}\label{fig:TEF_result_d}}
 \caption{(a),(b): An input video frame and its input motion vector field; (c),(d): TEF results of Algorithm~\ref{algorithm:coarse_to_fine} after 1 and 3 iterations, respectively ($T_{max}$=5 and $T_{step}$=1).}\label{fig:TEF_result}
\end{figure}

The long-term motion vector field with a large frame interval $T_{max}$ is first calculated and used to create the thermal energy field. Then, the TEF is iteratively updated with shorter-term motion vector fields, i.e., $\mathbf{F}_{\mathbf{P}}(T)$ with smaller $T$. Figs~\ref{fig:TEF_result_a} to~\ref{fig:TEF_result_d} show the TEF results after different iteration numbers. When more iterations are performed, more motion information with different intervals will be included in the thermal diffusion process. Thus, more precise results can be achieved in the TEF, as in Fig.~\ref{fig:TEF_result_d}. Fig.~\ref{fig:coherent_example_c} shows another TEF result after the entire coarse-to-fine thermal diffusion scheme. We find that:
\begin{enumerate}
 \item TEF is an enhanced version of the input motion where particles' energy directions in the TEF are similar to their original motion directions. Besides, since TEF include both the motion correlation among particles and the short-/long-term motion information among frames, coherent motions are effectively strengthened and highlighted in TEF.
 \item As mentioned, input motion vectors may be disordered, e.g., region A in Fig~\ref{fig:coherent_example_b}. However, the thermal energies from other particles can help recognize these disordered motion vectors and make them coherent, e.g., Fig.~\ref{fig:coherent_example_c}.
 \item Input motion vectors may be extremely small due to slow motion or occlusion by other objects (region B in Fig.~\ref{fig:coherent_example_b} and region C in Fig.~\ref{fig:TEF_result_b}). It is very difficult to include these particles into the coherent region by traditional methods  \cite{1,2,3,7} because they are close to the background motion vector. However, TEF can strengthen these small motion vectors by diffusing thermal energies from distant particles with larger motions.
\end{enumerate}

\subsection{Finding Coherent Motions through Triangulation}

Coherent motion regions can be achieved by performing segmentation on the TEF. We propose a triangulation-based scheme as follows:

{\bf Step 1: Triangulation.} In this step, we randomly sample particles from the entire scene and apply the triangulation  process \cite{19} to link the sampled particles. The block labeled as ``triangulation'' in Fig.~\ref{fig:framework} shows one triangulation result, where red dots are the sampled particles and the lines are links created by the triangulation process \cite{19}.

{\bf Step 2: Boundary detection.} We first obtain each triangulation link weight by:
\begin{equation}
 \omega\left( \mathbf{P},\mathbf{Q} \right)=\dfrac{||\mathbf{E}_{\mathbf{P}}-\mathbf{E}_{\mathbf{Q}}||}{||\mathbf{P}-\mathbf{Q}||}\label{equation:eq5}
\end{equation}
where $\mathbf{P}$ and $\mathbf{Q}$ are two connected particles, $\mathbf{E}_{\mathbf{P}}$ and $\mathbf{E}_{\mathbf{Q}}$ are the thermal energy vectors of $\mathbf{P}$ and $\mathbf{Q}$ in the TEF. A large weight will be assigned if the connected particles are from different coherent motion regions (i.e., they have different thermal energy vectors). Thus, by thresholding on the link weights, we can find links crossing the boundaries. The block labeled as ``detected region boundary'' in Fig.~\ref{fig:framework} shows one boundary detection result after step 2.

{\bf Step 3: Coherent motion segmentation.} Then, coherent motions can be easily segmented and we use the watershed algorithm \cite{35}. The final coherent motions are shown in the block named ``detected coherent motions'' in Fig.~\ref{fig:framework}.

\section{Constructing Semantic Regions\label{section:two_step_clustering}}

With the extracted coherent motions, accurate motion information in a frame can be achieved. However, since coherent motions vary over time, it is essential to construct semantic regions from time-varying coherent motions to catch stable semantic patterns inside a scene. For this purpose, we propose a \emph{two-step clustering scheme}. Assuming that in total $M$ coherent motions ($\mathbf{C}_m$, $m=1,..., M$) from $N$ TEFs extracted at $N$ times, the two-step clustering scheme is:

{\bf Step 1: Cluster coherent motion regions.} The similarity between two coherent motions $\mathbf{C}_m$ and $\mathbf{C}_k$ is computed as:
\begin{align}
 S_C(\mathbf{C}_m,\mathbf{C}_k)=\#\{&(\mathbf{P},\mathbf{Q})|\mathbf{P}\in\mathbf{L}_m, \mathbf{Q}\in\mathbf{L}_k, \nonumber\\ &\cos(\mathbf{E}_{\mathbf{P}},\mathbf{E}_{\mathbf{Q}})\cdot e^{-k_{p}||\mathbf{P}-\mathbf{Q}||^2}>\theta_{bp}\}\label{equation:eq6}
\end{align}
where $\#$\{$\cdot$\} is the number of elements in a set. $\theta_{bp}$ is a threshold which is set to be the same as $\theta_{c}$ in Eq.~\ref{equation:eq4} in our experiments. Furthermore, $\mathbf{L}_m$ and $\mathbf{L}_k$ are the sets of ``indicative particles'' for $\mathbf{C}_m$ and $\mathbf{C}_k$:
{\small
\begin{align}
 & \mathbf{L}_m=\{\mathbf{P}|\cos(\mathbf{E}_{\mathbf{P}},\mathbf{V}_{\mathbf{P}})>\theta_c, \mathbf{P} \text{ is on the boundary of $\mathbf{C}_m$}\} \nonumber \\
 & \mathbf{L}_k=\{\mathbf{Q}|\cos(\mathbf{E}_{\mathbf{Q}},\mathbf{V}_{\mathbf{Q}})>\theta_c, \mathbf{Q} \text{ is on the boundary of $\mathbf{C}_k$}\}\label{equation:eq7}
\end{align}}where $\mathbf{V}_{\mathbf{P}}=[v^x_{\mathbf{P}},v^y_{\mathbf{P}}]$ is the outer normal vector at $\mathbf{P}$, i.e., perpendicular to the boundary and pointing outward the coherent motion region, $\theta_c$ is the same threshold as in the condition for Eq.~\ref{equation:eq4}. That is, only particles which are on the boundaries of the coherent motion region and whose thermal energy vectors sharply point outward the region are selected as the indicative particles. Thus, we can avoid noisy particles and substantially reduce the required computations.

From Eq.~\ref{equation:eq6}, we can see that we first extract the indicative particles, then only utilize those high-correlation pairs, and the total number of such pairs are the similarity value between two coherent motions. It should be noted that the similarity will be calculated between any coherent motion pairs even if they belong to different TEFs.

Then, we construct a similarity graph for the $M$ coherent motions, and perform clustering \cite{24} on this similarity graph with the optimal number of clusters being determined automatically, the cluster results are grouped coherent regions.

\begin{figure}
 \centering
 \subfloat[]{\includegraphics[width=1.0\linewidth]{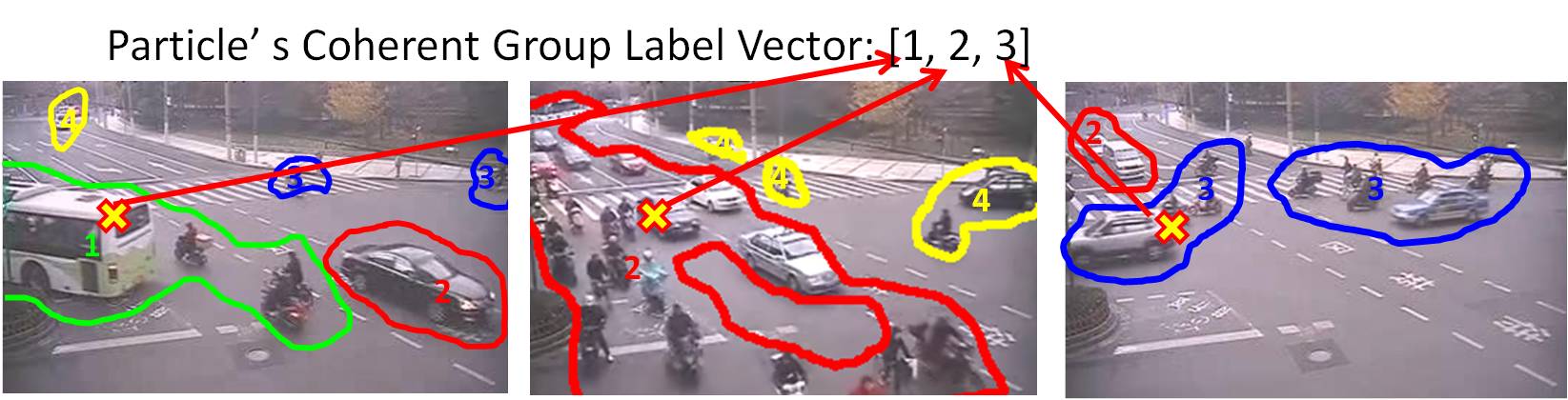}\label{fig:clustering_a}} \\
 \subfloat[]{\includegraphics[width=0.32\linewidth]{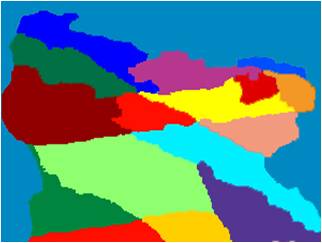}\label{fig:clustering_b}}
 \caption{(a) Step 1: Coherent regions in the three TEFs have been assigned different cluster labels by Step 1 and are displayed in different colors); (b) Find semantic regions by clustering the cluster label vectors of the particles (best viewed in color).}\label{fig:clustering}
\end{figure}

\begin{figure}
 \centering
  \subfloat[]{\includegraphics[width=0.28\linewidth]{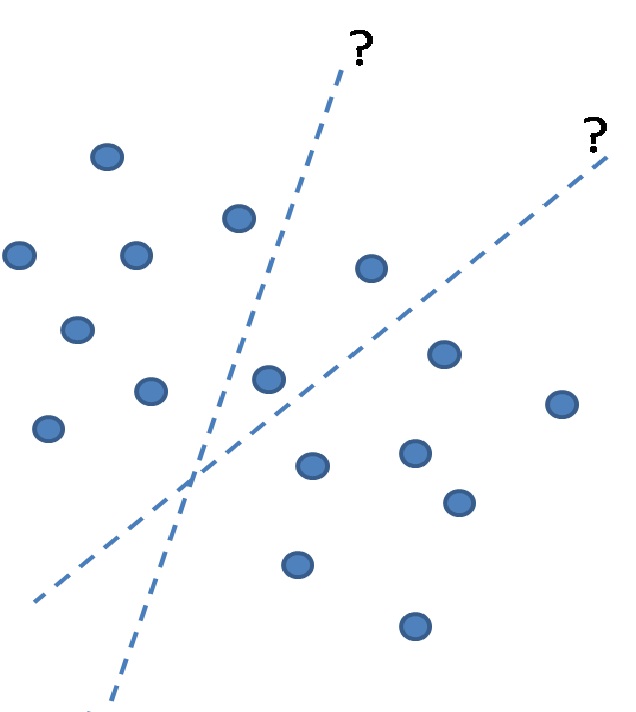} \label{fig:semantic_cluster_a}}
  \hspace{3pt}
  \subfloat[]{\includegraphics[width=0.32\linewidth,height=0.28\linewidth]{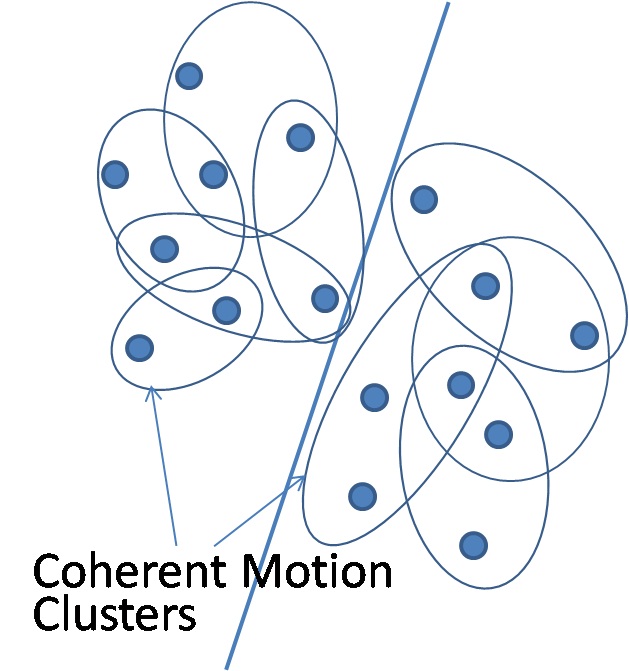} \label{fig:semantic_cluster_b}}
 \caption{(a) Directly segmenting semantic regions according to the particles' local features. (b) Segmenting semantic regions with the guidance of coherent motion clusters.} \label{fig:semantic_cluster}
\end{figure}

{\bf Step 2: Cluster to find semantic regions.} Each coherent motion is assigned a cluster label in Step 1, as illustrated in Fig.~\ref{fig:clustering_a}. However, due to the variation of coherent motions at different times, there exist many ambiguous particles. For example, in Fig.~\ref{fig:clustering_a}, the yellow cross particle belongs to different coherent motion clusters in different TEFs. This makes it difficult to directly use the clustered coherent motion results to construct reliable semantic regions. In order to address this problem, we further propose to encode particles in each TEF by the cluster labels of the particles' affiliated coherent motions. And by concatenating the cluster labels over different TEFs, we can construct a ``cluster label'' vector for each particle, as in Fig.~\ref{fig:clustering_a} And with these label vectors, the same spectral clustering process as Step 1 can be performed on the particles to achieve the final semantic regions, as in Fig.~\ref{fig:clustering_b}.

Comparing with previous semantic region segmentation methods \cite{22,23} which perform clustering using local similarity among particles, our scheme utilizes the guidance from the global coherent motion clustering results to strengthen the correlations among particles. For example, in Fig.~\ref{fig:semantic_cluster_a}, when directly segmenting the particles by their local features, its accuracy may be limited due to similar distances among particles. However, by utilizing cluster labels to encode the particles, similarities among particles can be suitably enhanced by the global coherent cluster information, as in Fig.~\ref{fig:semantic_cluster_b}. Thus, more precise segmentation results can be achieved.

\subsection{Recognizing Pre-defined Activities\label{section:pre_defined_recognition}}
Based on the constructed semantic regions, we are able to recognize pre-defined activities (i.e., activities with labeled training data) in the scene. In this paper, we simply average the TEF vectors in each semantic region and concatenate these averaged TEF vectors as the final feature vector for describing the activity patterns in a TEF. Then, a linear support vector machine (SVM) \cite{25} is utilized to train and recognize pre-defined activities. Experimental results show that with accurate TEF and precise semantic regions, we can achieve satisfying results using this simple method.

\subsection{Merging Disconnected Coherent Motions}
Since TEF also includes long-distance correlations between distant particles, by performing our clustering scheme, we also have the advantage of effectively merging disconnected coherent motions, which may be caused by the occlusion from other objects or low density of the crowd. For examples, the two disconnected blue regions in the right-most figure in Fig.~\ref{fig:clustering_a} are merged into the same cluster by our approach. Note that this issue is not well studied in the existing coherent motion research.

\section{Mining Recurrent Activities\label{section:cluster_and_merge}}

With the extracted coherent motions and constructed semantic regions, crowd activities can be recognized by constructing and pre-labeling training data, as in Section~\ref{section:pre_defined_recognition}. However, since pre-defining or labeling crowd activities take lots of human labors, it is also desirable to automatically mine recurrent activity patterns in a crowd scene without human intervention. For this purpose, we propose a \emph{cluster-and-marge process} which includes three steps: frame-level clustering, coherent motion merging, and flow curve extraction.

\subsection{Frame-level Clustering}
The frame-level clustering step clusters frames according to the extracted coherent motions and semantic regions, such that frames with the same recurrent activity pattern can be organized into the same group. In this paper, we first calculate inter-frame similarities for all frame pairs and then utilize spectral clustering \cite{24} to cluster frames according to these inter-frame similarities.


In order to calculate the inter-frame similarity between frames $t$ and $t-\tau$, the similarities between all coherent motions from frames $t$ and $t-\tau$ are first calculated using Eq.~\ref{equation:eq6}. Then, the inter-frame similarity $S_F\left(t,t- \tau\right)$ can be achieved from these coherent motion similarities and the segmented semantic regions. More specifically, we define $S_F\left(t,t- \tau\right)$ as
\begin{equation}
 S_F\left(t,t- \tau\right)=S_{FU}\left(t,t-\tau\right) \cdot S_{FM}\left(t,t-\tau\right)\label{equation:eq8}
\end{equation}
where $S_{FM}\left(t,t-\tau\right)$ is the similarity for the matched coherent motion pairs between $t$ and $t-\tau$, $S_{FU}\left(t,t-\tau\right)$ is the similarity for the unmatched coherent motion regions in frames $t$ and $t-\tau$. $S_{FM}\left(t,t-\tau\right)$ and $S_{FU}\left(t,t-\tau\right)$ can be calculated by Eqs~\ref{equation:eq9} and~\ref{equation:eq10}.

\begin{figure}
 \centering
 \includegraphics[width=0.55\textwidth]{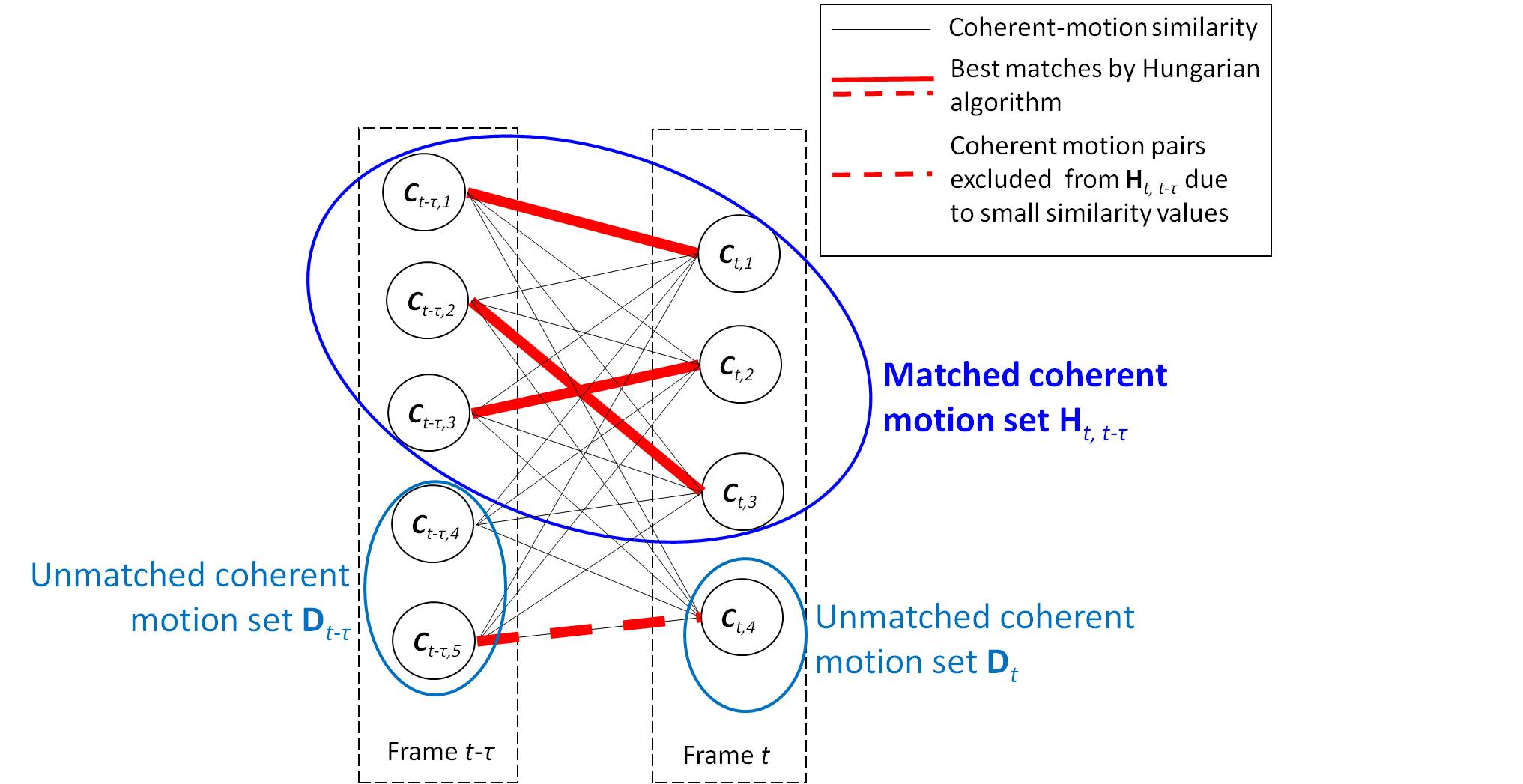}
 \caption{Example of matched and unmatched coherent motion region sets (best viewed in color).}\label{fig:matching}
\end{figure}

First, $S_{FM}\left(t,t-\tau\right)$ is defined as
{\small
\begin{equation}
 S_{FM}\left(t,t-\tau\right)=\frac{\sum\limits_{\left(\mathbf{C}_{t,i},\mathbf{C}_{t-\tau,j}\right)\in\mathbf{H}_{t,t-\tau}}\lambda_{i,j}S_C\left(\mathbf{C}_{t,i},\mathbf{C}_{t-\tau,j}\right)} {\max\{n_t,n_{t-\tau}\}}\label{equation:eq9}
\end{equation}}where $S_C\left(\mathbf{C}_{t,i},\mathbf{C}_{t-\tau,j}\right)$ is the similarity between coherent motion regions $\mathbf{C}_{t,i}$ and $\mathbf{C}_{t-\tau,j}$, $\lambda_{i,j}$ is the corresponding weight. $n_t$ and $n_{t-\tau}$ are the total number of coherent motion regions in frames $t$ and $t-\tau$, respectively. $\mathbf{H}_{t,t-\tau}$ is the set of all matched coherent region pairs. In this paper, $\mathbf{H}_{t,t-\tau}$ and $\lambda_{i,j}$ are calculated by the Hungarian algorithm \cite{34} which can achieve optimal coherent motion matching results based on the input coherent motion similarities. Furthermore, in order to exclude dissimilar coherent motion pairs from the matching result, coherent motion pairs $\left(\mathbf{C}_{t,i},\mathbf{C}_{t-\tau,j}\right)$ with small similarity values $S_C\left(\mathbf{C}_{t,i},\mathbf{C}_{t-\tau,j}\right)$ will be deleted from $\mathbf{H}_{t,t-\tau}$. Fig.~\ref{fig:matching} shows an example of the matched coherent motion pairs.

\begin{figure}
 \centering
 \includegraphics[width=0.55\linewidth]{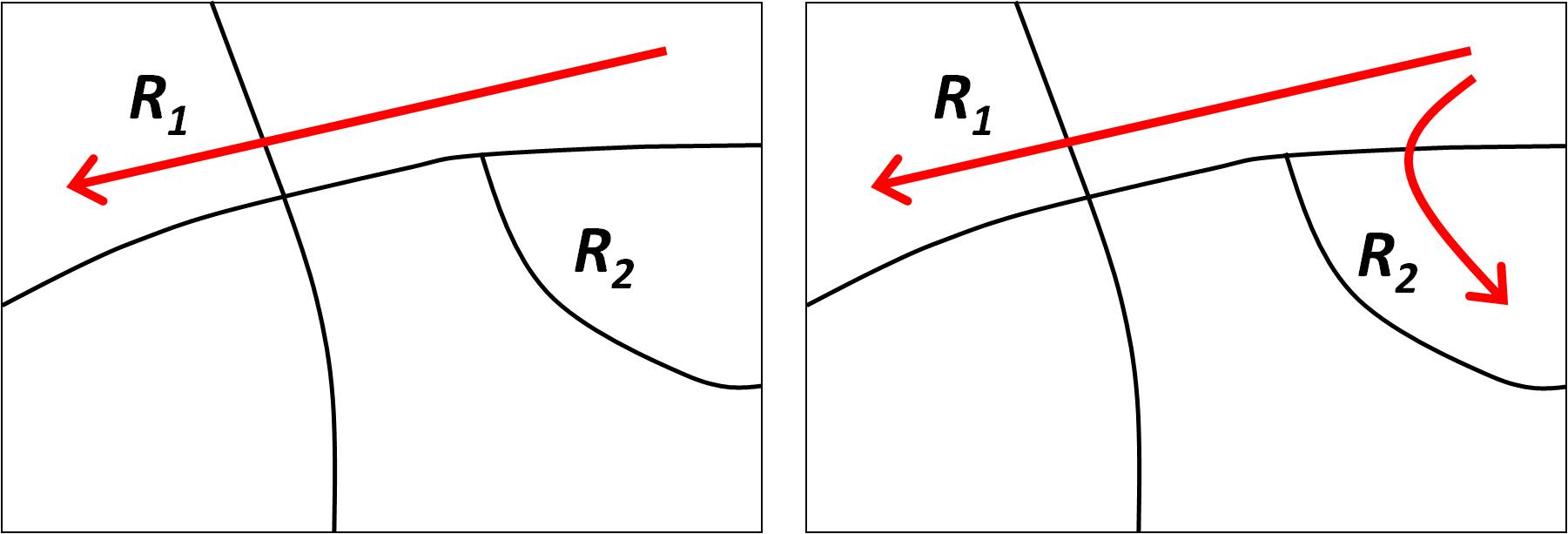}
 \caption{Motion flows for two recurrent activities displayed over semantic regions.}\label{fig:motion_flow_example}
\end{figure}

The next term $S_{FU}\left(t,t-\tau\right)$ is defined as
{\small
\begin{equation}
 S_{FU}\left(t,t-\tau\right)=\prod_{\mathbf{C}_{t-\tau,j}\in\mathbf{D}_{t-\tau}}\varepsilon\left(\mathbf{C}_{t-\tau,j}\right) \cdot \prod_{\mathbf{C}_{t,i}\in\mathbf{D}_{t}}\varepsilon\left(\mathbf{C}_{t,i}\right)\label{equation:eq10}
\end{equation}}where $\mathbf{D}_{t-\tau}$ and $\mathbf{D}_{t}$ are the sets of unmatched coherent regions in frames $t-\tau$ and $t$, as shown in Fig.~\ref{fig:matching}. $\varepsilon\left(\mathbf{C}\right)$ is the unmatching cost for coherent motion region $\mathbf{C}$:
\begin{equation}
\varepsilon\left(\mathbf{C}\right)=\dfrac {\sum_{\mathbf{R}_k,\mathbf{R}_k\cap\mathbf{C}\neq\emptyset}{\dfrac{1}{\rho\left(\mathbf{R}_k\right)}}}{\#\{\mathbf{R}_k|\mathbf{R}_k\cap\mathbf{C}
\neq\emptyset\}}\label{equation:eq11}
\end{equation}
where $\mathbf{R}_k$ is the $k$-th semantic region of the scene, the term $\#\{\mathbf{R}_k|\mathbf{R}_k\cap\mathbf{C}
\neq\emptyset\}$ represents the total number of semantic regions that have overlap with the coherent motion region $\mathbf{C}$. $\rho\left(\mathbf{R}_k\right)$ is the importance cost measuring whether semantic region $\mathbf{R}_k$ is important in distinguishing different recurrent activities. For example, assuming that a scene includes two recurrent activities, as in Fig.~\ref{fig:motion_flow_example}, it is obvious that the semantic region $\mathbf{R}_2$ on the right should have larger importance cost since the two recurrent activity patterns have different motion flows in $\mathbf{R}_2$. Comparatively, the semantic region $\mathbf{R}_1$ on the left should have smaller importance cost since both recurrent activity patterns have similar flows in $\mathbf{R}_1$. Therefore, when calculating the similarity between frames $t$ and $t-\tau$, if there exists an unmatched coherent region $\mathbf{C}$ in $\mathbf{R}_2$, a large importance cost $\rho\left(\mathbf{R}_2\right)$ will be applied to reduce the inter-frame similarity, indicating that frames $t$ and $t-\tau$ have different recurrent activity patterns. On the contrary, if there exists an unmatched coherent region $\mathbf{C}$ in $\mathbf{R}_1$, the inter-frame similarity will be less affected since a coherent region in $\mathbf{R}_1$ is less indicative of the differences between recurrent activities.

For $\rho\left(\mathbf{R}_k\right)$, we first perform a pre-clustering according to the matched coherent motion similarities $S_{FM}\left(t,t-\tau\right)$ which roughly clusters frames into different recurrent activity groups. Then a vector is constructed for each semantic region $\mathbf{R}_k$:
$\left[Num_{\mathbf{R}_k,\mathbf{G}_1},Num_{\mathbf{R}_k,\mathbf{G}_2},...,Num_{\mathbf{R}_k,\mathbf{G}_Z}\right]$ where $Num_{\mathbf{R}_k,\mathbf{G}_i}$ is the total number of coherent motions located in $\mathbf{R}_k$ in the $i$-th pre-clustered recurrent activity group $\mathbf{G}_i$, $Z$ is the total number of pre-clustered recurrent activity groups. Finally, $\rho\left(\mathbf{R}_k\right)$ can be calculated by:
\begin{equation}
\rho\left(\mathbf{R}_k\right)=e^{k_s \cdot var\left\{Num_{\mathbf{R}_k,\mathbf{G}_1},Num_{\mathbf{R}_k,\mathbf{G}_2},...,Num_{\mathbf{R}_k,\mathbf{G}_Z}\right\}}\label{equation:eq12}
\end{equation}
where $var\{\cdot\}$ is the variance operation, $k_s= \frac{1}{{Num_{f}}^2}$ where $Num_{f}$ is the total number of frames to be clustered. According to Eq.~\ref{equation:eq12}, if coherent motions appear evenly in $\mathbf{R}_k$ for different recurrent activities, i.e., the variance is smaller, it implies that $\mathbf{R}_k$ is less important in distinguishing different recurrent activities. On the contrary, if the appearance time of coherent motions in $\mathbf{R}_k$ has larger variation over different pre-clustered recurrent activity groups, a large $\rho\left(\mathbf{R}_k\right)$ will be achieved to increase the importance of $\mathbf{R}_k$. The complete process of frame-level clustering is illustrated in Algorithm~\ref{algorithm:frame_clustering}.

\begin{algorithm}
 \caption{Frame-level Clustering Process}
 \small{
   \begin{adjustwidth}{6pt}{0pt}
   {\bf Input}: Coherent regions $\mathbf{C}_{t,i}$ extracted for each frame $t$, and semantic regions $\mathbf{R}_k$ of the scene\\
   \hspace{9pt} {\bf Output}: Recurrent activity groups including frames with similar recurrent activity patterns
   \end{adjustwidth}
   }

 \begin{algorithmic}[1]
 \small{
   \STATE Calculate similarities $S_C\left(\mathbf{C}_{t,i},\mathbf{C}_{t-\tau,j}\right)$ between all coherent motion regions from different frames
   \STATE Calculate $S_{FM}\left(t,t-\tau\right)$ for all frame pairs based on $S_C\left(\mathbf{C}_{t,i},\mathbf{C}_{t-\tau,j}\right)$
   \STATE Pre-cluster frames based on $S_{FM}\left(t,t-\tau\right)$
   \STATE Calculate importance cost $\rho\left(\mathbf{R}_k\right)$ for all semantic regions based on the pre-clustering result
   \STATE Calculate $S_{FU}\left(t,t-\tau\right)$ for all frame pairs according to the unmatched coherent regions and $\rho\left(\mathbf{R}_k\right)$
   \STATE Calculate inter-frame similarities $S_{F}\left(t,t-\tau\right)$ for all frame pairs using $S_{FM}\left(t,t-\tau\right)$ and $S_{FU}\left(t,t-\tau\right)$
   \STATE With $S_{F}\left(t,t-\tau\right)$, cluster frames into recurrent activity groups
   \STATE Output the clustering result in line 9
 }
 \end{algorithmic}\label{algorithm:frame_clustering}
\end{algorithm}

\subsection{Coherent Motion Merging}

\begin{figure}
 \centering
 \includegraphics[width=0.75\linewidth]{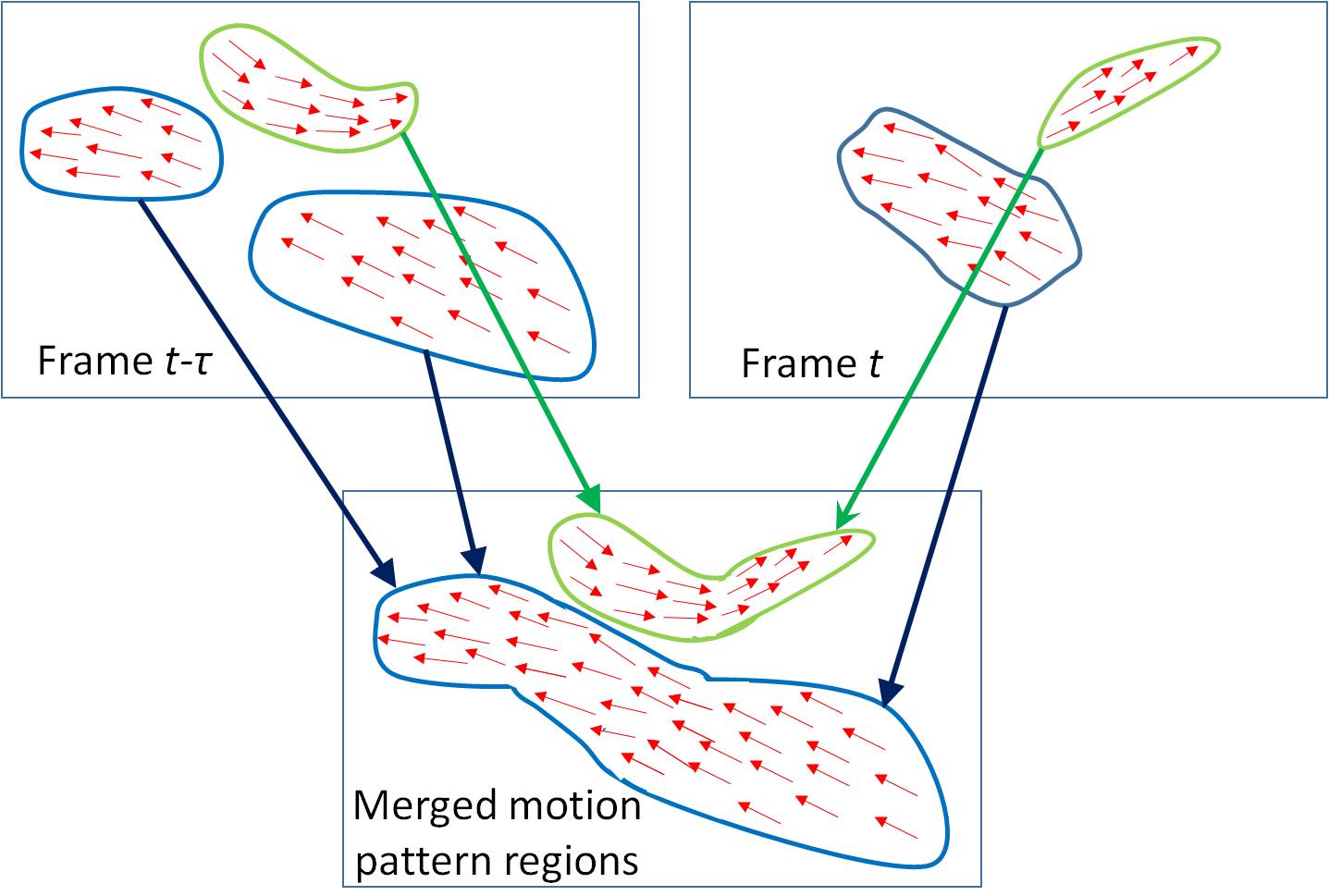}
 \caption{Process of similar coherent motion merging. Frames $t$ and $t-\tau$ are from the same recurrent activity group, the green coherent motion regions in frames $t$ and $t-\tau$ belong to one coherent motion cluster, and the blue coherent motion regions in frames $t$ and $t-\tau$ belong to another coherent motion cluster. (Best viewed in color.)}\label{fig:coherent_merging}
\end{figure}

After frame-level clustering, frames are clustered into different recurrent activity groups. Thus, by parsing frames in each recurrent activity group, complete motion patterns for each recurrent activity can be estimated. In this paper, we introduce a coherent motion merging step to merge similar coherent motions from the same recurrent activity group for achieving motion pattern regions. More specifically, we first apply the same operation as Step 1 in the two-step clustering scheme (Section~\ref{section:two_step_clustering}) to cluster coherent motion regions inside the same recurrent activity group. Then, coherent motions of the same cluster are merged together to form a motion pattern region. The merging process can be described by Eq.~\ref{equation:eq13} and Fig.~\ref{fig:coherent_merging}.
\begin{equation}
 \overline{\mathbf{E}}_{\mathbf{P},\mathbf{\Psi}_j} = \sum_{\mathbf{C}_m\in\mathbf{\Psi}_j}{\dfrac{\mathbf{E}_{\mathbf{P},\mathbf{C}_m}}{\#\{\mathbf{\Psi}_j\}}}\label{equation:eq13}
\end{equation}
if $\dfrac{\#\left\{\mathbf{E}_{\mathbf{P},\mathbf{C}_m}|\mathbf{E}_{\mathbf{P},\mathbf{C}_m}\neq[0,0],\mathbf{C}_m\in\mathbf{\Psi}_j\right\}}{\#\{\mathbf{\Psi}_j\}}>\theta_{mf}$, and it is $\left[0,0\right]$ if otherwise,
where $\mathbf{\Psi}_j$ the $j$-th coherent motion cluster. $\overline{\mathbf{E}}_{\mathbf{P},\mathbf{\Psi}_j}=\left[\overline{E}^x_{\mathbf{P},\mathbf{\Psi}_j},\overline{E}^y_{\mathbf{P},\mathbf{\Psi}_j}\right]$ is the merged motion vector result for $\mathbf{\Psi}_j$ at particle $\mathbf{P}$. $\mathbf{C}_m$ is a coherent motion region belonging to coherent motion cluster $\mathbf{\Psi}_j$. $\#\{\mathbf{\Psi}_j\}$ is the total number of coherent motion regions in cluster $\mathbf{\Psi}_j$. $\theta_{mf}$ is a threshold which is set as 0.4 in our experiments. $\mathbf{E}_{\mathbf{P},\mathbf{C}_m}$ is the TEF thermal energy for $\mathbf{C}_m$ at particle $\mathbf{P}$. Note that $\mathbf{E}_{\mathbf{P},\mathbf{C}_m}$ is set to $\left[0,0\right]$ if $\mathbf{P}$ is outside the region of $\mathbf{C}_m$. And $\#\left\{\mathbf{E}_{\mathbf{P},\mathbf{C}_m}|\mathbf{E}_{\mathbf{P},\mathbf{C}_m}\neq[0,0],\mathbf{C}_m\in\mathbf{\Psi}_j\right\}$ is the total number of non-zero TEF thermal energies at particle $\mathbf{P}$ and belonging to $\mathbf{\Psi}_j$.

According to Eq.~\ref{equation:eq13}, the merged motion pattern region $\overline{\mathbf{R}}_{\mathbf{\Psi}_j}=\{\overline{\mathbf{E}}_{\mathbf{P},\mathbf{\Psi}_j}\}$ for a coherent motion cluster $\mathbf{\Psi}_j$ is basically the normalized summation over all coherent regions in $\mathbf{\Psi}_j$. Besides, we further introduce a threshold $\theta_{mf}$ to filter out noisy or isolated particles which have low frequent motions in the coherent motion cluster $\mathbf{\Psi}_j$. An example of merged motion pattern regions is shown in Fig.~\ref{fig:coherent_merging}.

\subsection{Flow Curve Extraction}

The motion pattern regions achieved in the previous step can represent the complete motion information for each recurrent activity. However, since motion pattern regions may overlap with each other and the contours of motion pattern regions may also be irregular, it is necessary to extract flow curves from these motion pattern regions such that recurrent activities can be more clearly described and visualized.

\begin{figure}
 \centering
 \includegraphics[width=0.4\textwidth]{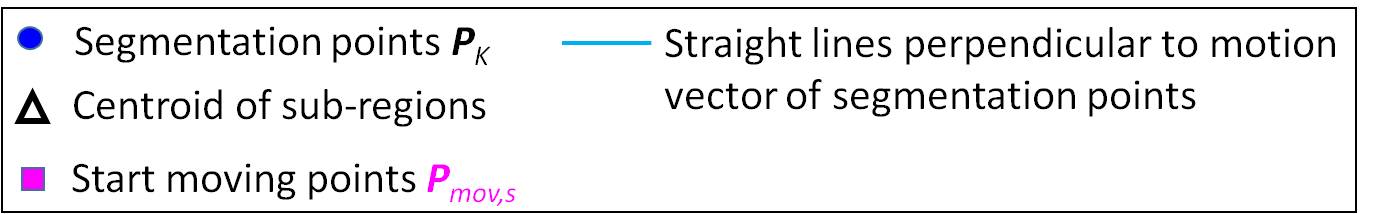}\\
 \centering
 \subfloat[]{\includegraphics[width=0.55\linewidth]{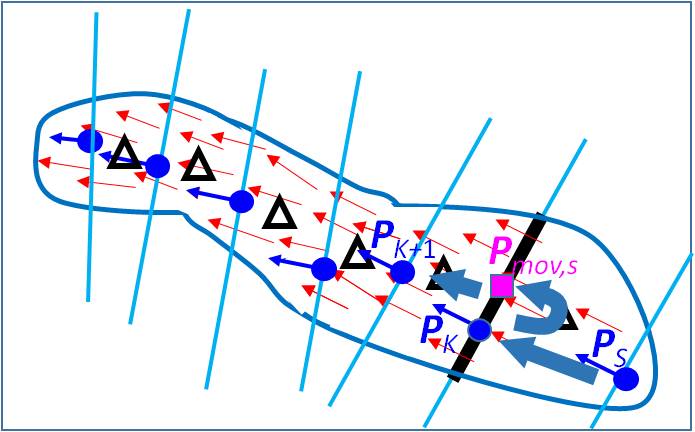}\label{fig:curve_extraction_a}}\\
 \subfloat[]{\includegraphics[width=0.55\linewidth]{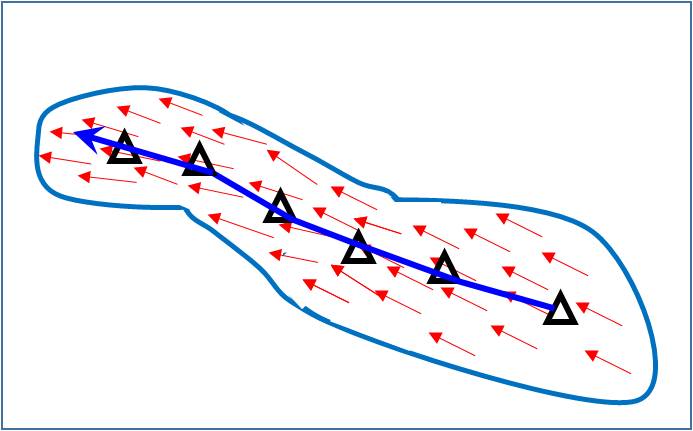}\label{fig:curve_extraction_b}}\\
 \caption{Flow curve extraction process. Finding segmentation points, draw straight lines to achieve sub-regions, and calculate centroids of each sub-region; (b) Link centroids to achieve the extracted flow curve. (Best viewed in color.)}\label{fig:curve_extraction}
\end{figure}

Our proposed flow curve extraction process can be described by Algorithm~\ref{algorithm:flow_curve} and Fig.~\ref{fig:curve_extraction}. According to Algorithm~\ref{algorithm:flow_curve} and Fig.~\ref{fig:curve_extraction}, our approach first sequentially cuts a motion pattern region $\overline{\mathbf{R}}_{\mathbf{\Psi}_j}$ into sub-regions along the motion direction in $\overline{\mathbf{R}}_{\mathbf{\Psi}_j}$. Then the centroids of sub-regions are linked together to achieve the output flow curve. With the above process, the extracted flow curve can accurately catch the major motion flow of a motion pattern region. Furthermore, it should be noted that in step 5 of Algorithm~\ref{algorithm:flow_curve}, if the line perpendicular to the motion vector $\overline{\mathbf{E}}_{\mathbf{P}_{K+1},\mathbf{\Psi}_j}$ at $\mathbf{P}_{K+1}$ is intersecting with a branched motion region (i.e., the motion region diverges around $\mathbf{P}_{K+1}$), multiple $\mathbf{P}_{mov,s}$ points will be achieved and the following flow curve extraction process will be performed on each $\mathbf{P}_{mov,s}$ respectively. In this way, we can properly achieve branched flow curves at the branch region.

\begin{algorithm}
 \caption{Flow Curve Extraction}
 \small{
   \begin{adjustwidth}{5pt}{0pt}
   {\bf Input}: A motion pattern region $\overline{\mathbf{R}}_{\mathbf{\Psi}_j}$ merged from coherent region cluster $\mathbf{\Psi}_j$\\
   {\bf Output}: A flow curve extracted from $\overline{\mathbf{R}}_{\mathbf{\Psi}_j}$
   \end{adjustwidth}
 }
 \begin{algorithmic}[1]
 \small{
  \STATE Calculate the skeleton of $\overline{\mathbf{R}}_{\mathbf{\Psi}_j}$ \cite{35}
  \STATE Find the end point $\mathbf{P}_s$ of the skeleton which is on ``backward'' position to all other end points, where the ``backward'' direction is defined as the reversed direction of the motion flows in $\overline{\mathbf{R}}_{\mathbf{\Psi}_j}$
  \STATE $\mathbf{P}_K=\mathbf{P}_s$, where $\mathbf{P}_K$ is the current segmentation point
  \WHILE{$\mathbf{P}_{K+1}$ is inside $\overline{\mathbf{R}}_{\mathbf{\Psi}_j}$}
    \STATE $\mathbf{P}_{mov,s}$ as the middle point of the line perpendicular to the motion vector $\overline{\mathbf{E}}_{\mathbf{P}_{K},\mathbf{\Psi}_j}$ at $\mathbf{P}_K$
    \FOR{$n$=0 to $Num_{mov}$  \{$Num_{mov}$ is the number of movements\}}
      \STATE Move from $\mathbf{P}_{mov,s}$ to $\mathbf{P}_{mov,e}$ by $\overline{\mathbf{E}}_{\mathbf{P}_{mov,s},\mathbf{\Psi}_j}$, where $\overline{\mathbf{E}}_{\mathbf{P}_{mov,s},\mathbf{\Psi}_j}$ is motion vector at $\mathbf{P}_{mov,s}$ in $\overline{\mathbf{R}}_{\mathbf{\Psi}_j}$
      \STATE $\mathbf{P}_{mov,s}$=$\mathbf{P}_{mov,e}$
    \ENDFOR
    \STATE $\mathbf{P}_{K+1}$= $\mathbf{P}_{mov,s}$, where $\mathbf{P}_{K+1}$ is the next segmentation point
    \STATE Draw two straight lines perpendicular to the motion vectors of at $\mathbf{P}_K$ and $\mathbf{P}_{K+1}$, respectively
    \STATE Calculate the centroid of the sub-region segmented by the lines in line 13
    \STATE $\mathbf{P}_K=\mathbf{P}_{K+1}$
  \ENDWHILE
  \STATE Sequentially link together all centroid points achieved by line 14
  \STATE Smooth the linked curve by line 17
  \STATE Output the curve by line 18
 }
 \end{algorithmic}\label{algorithm:flow_curve}
\end{algorithm}

\section{Experimental Results\label{section:experiments}}

Our approach is implemented by Matlab and the optical flow fields  \cite{6} are used as the input motion vector fields while each pixel in the frame is viewed as a particle. In order to achieve motion vector fields with $T$-frame intervals ($T=10$ in our experiments), the particle advection method \cite{1} is used which tracks the movement of each particle over $T$ frames.

\subsection{Results for Coherent Motion Detection}

\begin{figure*}
 \centering
 \subfloat[]{\includegraphics[width=0.125\linewidth,height=0.5\linewidth]{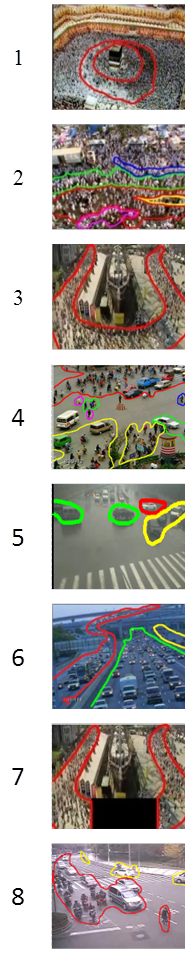}\label{fig:coherent_motion_results_a}}
 \hspace{2pt}
 \subfloat[]{\includegraphics[width=0.09\linewidth,height=0.5\linewidth]{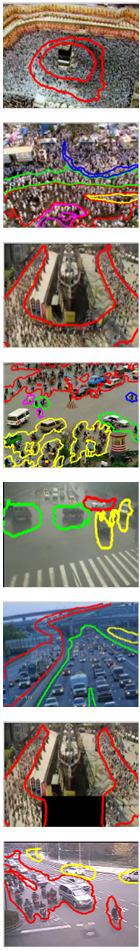}\label{fig:coherent_motion_results_b}}
 \hspace{2pt}
 \subfloat[]{\includegraphics[width=0.09\linewidth,height=0.5\linewidth]{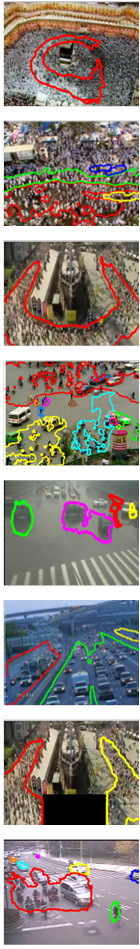}\label{fig:coherent_motion_results_c}}
 \hspace{2pt}
 \subfloat[]{\includegraphics[width=0.09\linewidth,height=0.5\linewidth]{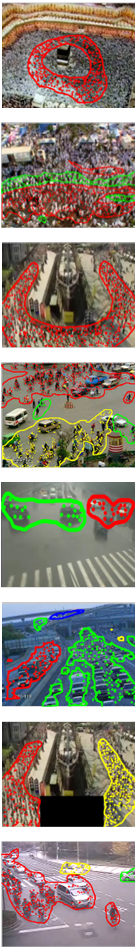}\label{fig:coherent_motion_results_d}}
 \hspace{2pt}
 \subfloat[]{\includegraphics[width=0.09\linewidth,height=0.5\linewidth]{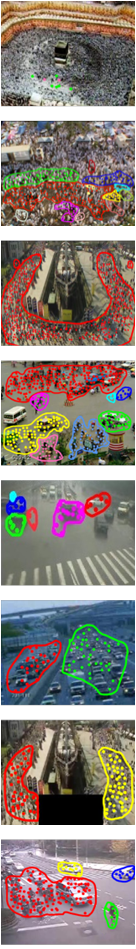}\label{fig:coherent_motion_results_e}}
 \hspace{2pt}
 \subfloat[]{\includegraphics[width=0.09\linewidth,height=0.5\linewidth]{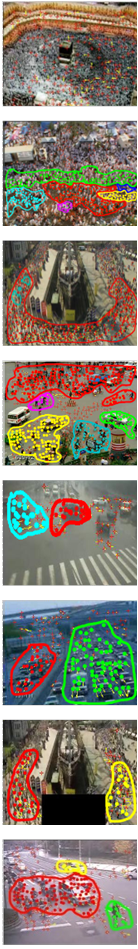}\label{fig:coherent_motion_results_f}}
 \hspace{2pt}
 \subfloat[]{\includegraphics[width=0.09\linewidth,height=0.5\linewidth]{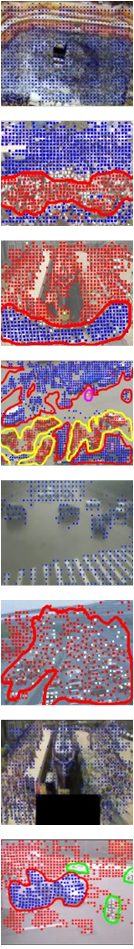}\label{fig:coherent_motion_results_g}}
 \hspace{2pt}
 \subfloat[]{\includegraphics[width=0.09\linewidth,height=0.5\linewidth]{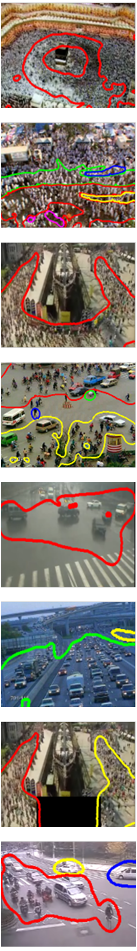}\label{fig:coherent_motion_results_h}}
 \caption{Coherent motion extraction results. (a): Ground Truth, (b): Results of our approach, (c): Results of \cite{1}, (d): Results of \cite{2}, (e): Results of \cite{3}, (f): Results of \cite{7}, (g): Results of \cite{12}, (h): Results of \cite{26}. (Best viewed in color)}\label{fig:coherent_motion_results}
\end{figure*}

We perform experiments on a dataset including 30 different crowd videos collected from the UCF dataset  \cite{1},  the UCSD dataset  \cite{16}, the CUHK dataset  \cite{7}, and our own collected set. This dataset covers various real-world crowd scene scenarios with both low- and high-density crowds and both rapid and slow motion flows. Some example frames of the dataset is shown in Fig.~\ref{fig:coherent_motion_results}.

We compare our approach with four state-of-the-art coherent motion detection algorithms: The Lagrangian particle dynamics approach \cite{1}, the local-translation domain segmentation approach  \cite{2}, the coherent-filtering approach \cite{3}, and the collectiveness measuring-based approach  \cite{7}. In order to further demonstrate the effectiveness of our approach, we also include the results of a general motion segmentation method \cite{12} and an anisotropic-diffusion-based image segmentation method \cite{26}.

{\bf Qualitative comparison on coherent motion detection.} Fig.~\ref{fig:coherent_motion_results} compares the coherent motion detection results for different methods. We include the manually labeled ground truth results in the first column. From Fig.~\ref{fig:coherent_motion_results}, we can see that our approach can achieve better coherent motion extraction than the compared methods. For example, in sequence 1, our approach can effectively extract the circle-shape coherent motion. Comparatively, the method in \cite{1} can only detect part of the circle while the methods in \cite{3} and \cite{7} fail to work since few reliable key points are extracted from this over-crowded scene. For sequences 2 and 4 where multiple complex motion flows exist, our approach can still precisely detect the small and less differentiable coherent motions, such as the pink region on the bottom and the blue region on the top in sequence 2~\subref{fig:coherent_motion_results_a}. The compared methods have low effectiveness in identifying these regions due to the interference from the neighboring motion regions. In sequences 3 and 6, since motions on the top of the frame are extremely small and close to the background, the compared methods fail to include these particles into the coherent motion region. However, in our approach, these small motions can be suitably strengthened and included through the thermal diffusion process. Furthermore, the methods in \cite{12} and \cite{26} do not show satisfying results, e.g., in sequences 5 and 6. This is because: (1) the crowd scenes are extremely complicated such that the extracted particle flows or trajectories become unreliable, thus making the general motion segmentation methods \cite{12} difficult to create precise results; (2) Since many coherent region boundaries in the crowd motion fields are rather vague and unrecognizable, good boundaries cannot be easily achieved without suitably utilizing the characteristics of the motion vector fields. Thus, simply applying the existing anisotropic-diffusion segmentation methods \cite{26} cannot achieve satisfying results.

{\bf Capability to handle disconnected coherent motions.} Sequences 5-8 in Fig.~\ref{fig:coherent_motion_results} compare the algorithms' capability in handling disconnected coherent motions. In sequence 7, we manually block one part of the coherent motion region while in sequences 5, 6, and 8, the red or green coherent motion regions are disconnected due to occlusion by other objects or low density. Since the disconnected regions are separated far from each other, most compared methods wrongly segment them into different coherent motion regions. However, with our thermal diffusion process and two-step clustering scheme, these regions can be successfully merged into one coherent region.

{\bf Quantitative comparison.} Table~\ref{table:table_CNE} compares the quantitative results for different methods. In Table~\ref{table:table_CNE}, the average Particle Error Rates (PERs) and the average Coherent Number Error (CNE) for all the sequences in our dataset are compared to measure the overall accuracy of coherent motion detection. PER is calculated by PER = \# of Wrong Particles $/$ Total \# of Particles. CNE is calculated by $CNE=\dfrac {\sum _{i}| Num_{d}(i) -Num_{gt}(i) |} {\Sigma _{i}1}$ where $Num_{d}(i)$ and $Num_{gt}(i)$ are the numbers of detected and ground-truth coherent regions for sequence $i$, respectively, $\Sigma _{i}1$ is the total number of sequences.

\begin{table}
 \caption{Average PER and CNE for all sequences in the dataset.} \label{comparison1}
 \small
 \centering
  \begin{tabular}{|@{\;}c@{\;}|@{\;}c@{\;}|@{\;}c@{\;}|@{\;}c@{\;}|@{\;}c@{\;}|@{\;}c@{\;}|@{\;}c@{\;}|@{\;}c@{\;}|}
    \hline
        &  Proposed &\cite{1}&\cite{2}&\cite{3}&\cite{7}&\cite{12}&\cite{26}\\
    \hline
       PER $(\%)$&{\bf 7.8}&32.5&19.5&25.6&24.1&66.4&21.4\\
    \hline
       CNE&{\bf 0.14}&1.24&0.93&1.05&0.96&1.78&0.84\\
    \hline
  \end{tabular}\label{table:table_CNE}
\end{table}

Table~\ref{table:table_CNE} further demonstrates the effectiveness of our approach. In Table~\ref{table:table_CNE}, we can see that 1) Our approach can achieve smaller coherent detection error rates than the other methods, 2) Our approach can accurately obtain the coherent region numbers (close to the ground truth) while other methods often over-segment or under-segment the coherent regions.

\begin{figure}
\centering
 \subfloat[$k_p=0.2$]{\includegraphics[width=0.185\linewidth]{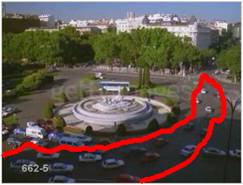}\label{fig:different_parameter_a}}
 \hspace{0.2pt}
 \subfloat[$k_p=0.5$]{\includegraphics[width=0.185\linewidth]{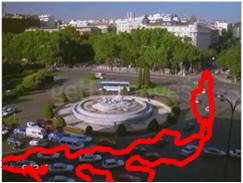}\label{fig:different_parameter_b}}
 \hspace{0.2pt}
 \subfloat[$k_p=0.7$]{\includegraphics[width=0.185\linewidth]{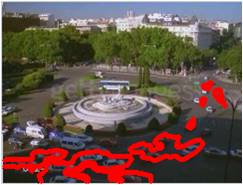}\label{fig:different_parameter_c}}
 \hspace{0.2pt}
 \subfloat[$k_f=0.6$]{\includegraphics[width=0.185\linewidth]{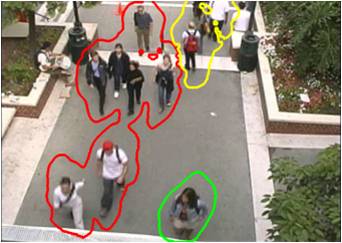}\label{fig:different_parameter_d}}
 \hspace{0.2pt}
 \subfloat[$k_f=0.9$]{\includegraphics[width=0.185\linewidth]{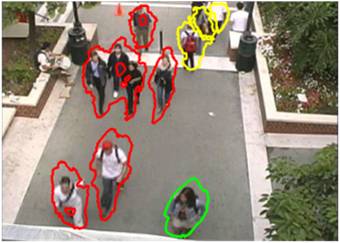}\label{fig:different_parameter_e}}
 \caption{The coherent motion detection results of our approach under different $k_p$ and $k_f$ values.}\label{fig:different_parameter}
\end{figure}

{\bf Effect of different parameter values.} Finally, Fig.~\ref{fig:different_parameter} shows the results of our approach under different parameter values, i.e., $k_p$ and $k_f$ in Eqs~\ref{equation:eq3} and~\ref{equation:eq4}. From Figs~\ref{fig:different_parameter_a} to~\ref{fig:different_parameter_c}, we can see that $k_p$ mainly governs the thermal diffusion distance. A small $k_p$ will make the thermal energies to be diffused farther and thus can achieve larger coherent motion regions. When $k_p$ increases, the extracted coherent motion region will shrink. Furthermore, $k_f$ determines the directivity of thermal diffusion. When $k_f$ increases, the diffused thermal energies will concentrate more along the motion direction of the source heat particles. On the contrary, when $k_f$ decreases, the thermal energies will be propagated more uniformly to all directions around the heat source particle. Thus, the boundaries will shrink horizontally with larger $k_f$, as in Fig.~\ref{fig:different_parameter_e}. However, note that in all examples in Fig.~\ref{fig:different_parameter}, our approach can always suitably merge coherent regions together even when they become disconnected when the parameter value changes.

\subsection{Results for Semantic Region Construction and Pre-defined Activity Recognition}

\begin{figure}
  \centering
  \subfloat[HD]{\includegraphics[width=0.2\linewidth,height=0.12\linewidth]{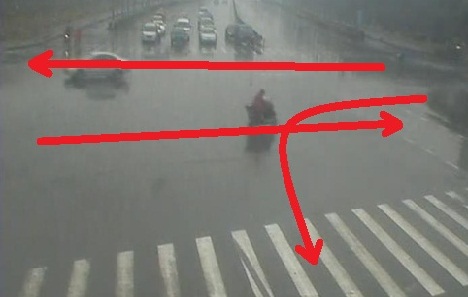}\label{fig:groundtruth_acitivity_a}}
  \hspace{0.1mm}
  \subfloat[HP]{\includegraphics[width=0.2\linewidth,height=0.12\linewidth]{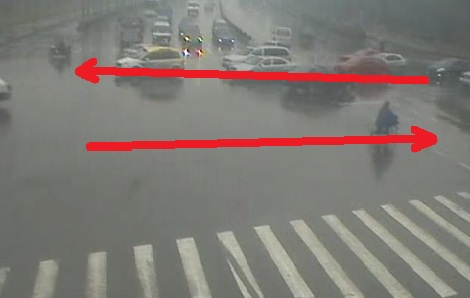}\label{fig:groundtruth_acitivity_b}}
  \hspace{0.1mm}
  \subfloat[BT]{\includegraphics[width=0.2\linewidth,,height=0.12\linewidth]{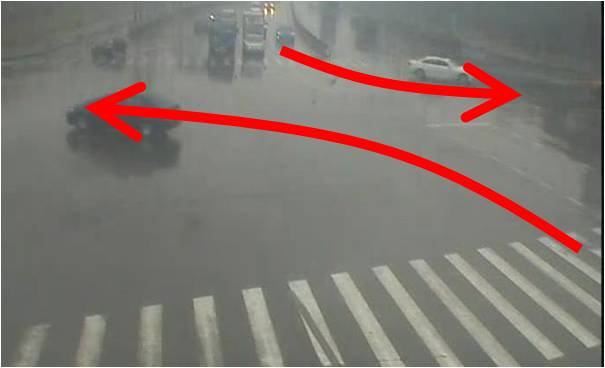}\label{fig:groundtruth_acitivity_c}}
  \hspace{0.1mm}
  \subfloat[VP]{\includegraphics[width=0.2\linewidth,height=0.12\linewidth]{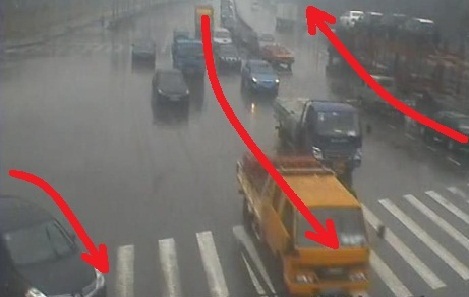}\label{fig:groundtruth_acitivity_d}}\\
  \subfloat[VL]{\includegraphics[width=0.2\linewidth,height=0.12\linewidth]{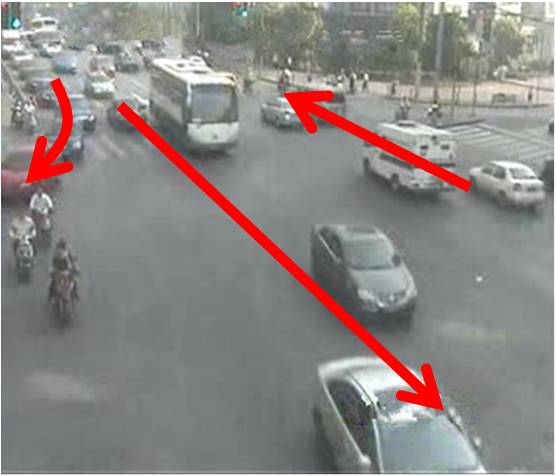}\label{fig:groundtruth_acitivity_e}}
  \hspace{0.1mm}
  \subfloat[BT]{\includegraphics[width=0.2\linewidth,height=0.12\linewidth]{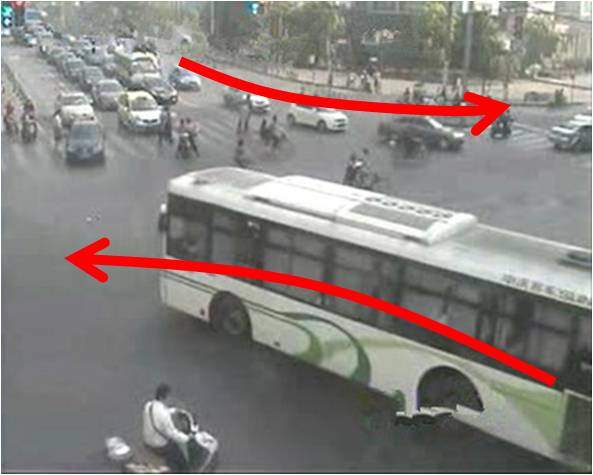}\label{fig:groundtruth_acitivity_f}}
  \hspace{0.1mm}
  \subfloat[HP]{\includegraphics[width=0.2\linewidth,height=0.12\linewidth]{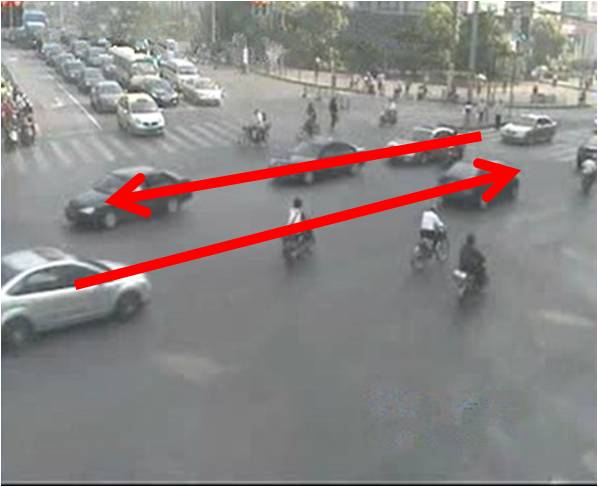}\label{fig:groundtruth_acitivity_g}}
  \hspace{0.1mm}
  \subfloat[HU]{\includegraphics[width=0.2\linewidth,height=0.12\linewidth]{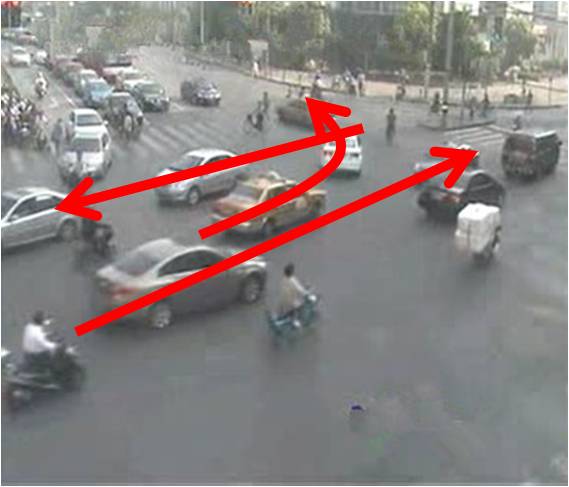}\label{fig:groundtruth_acitivity_h}}\\
  \subfloat[VR]{\includegraphics[width=0.185\linewidth,height=0.12\linewidth]{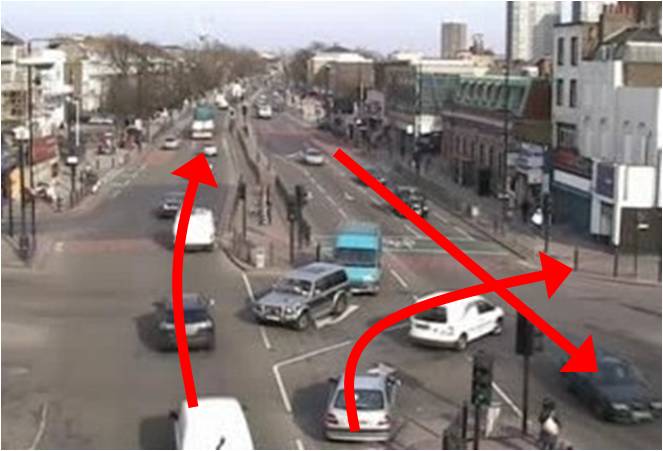}\label{fig:groundtruth_acitivity_i}}
  \hspace{0.1mm}
  \subfloat[HU]{\includegraphics[width=0.185\linewidth,height=0.12\linewidth]{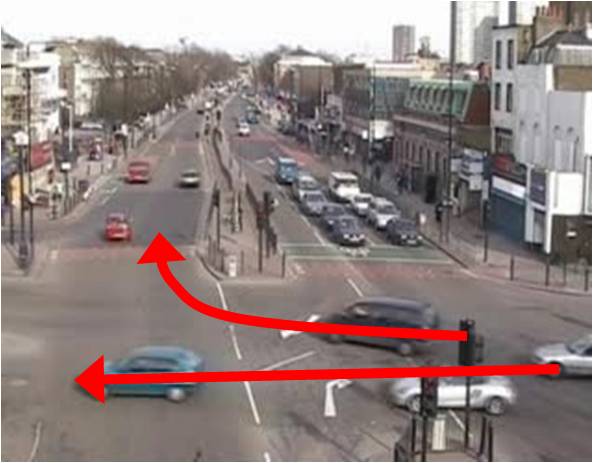}\label{fig:groundtruth_acitivity_j}}
  \hspace{0.1mm}
  \subfloat[VP]{\includegraphics[width=0.185\linewidth,height=0.12\linewidth]{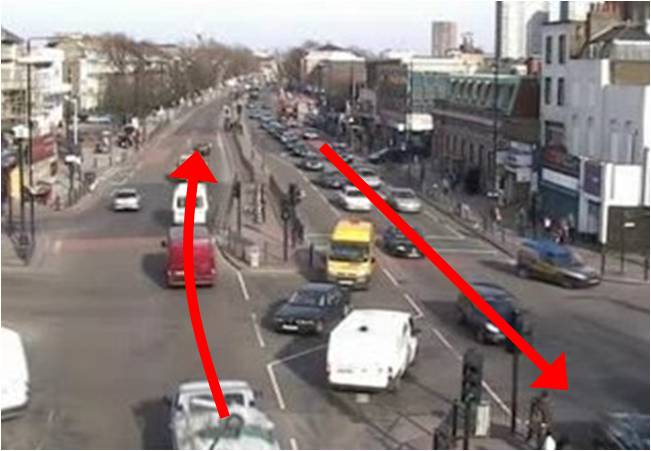}\label{fig:groundtruth_acitivity_k}}
  \hspace{0.1mm}
  \subfloat[HD]{\includegraphics[width=0.185\linewidth,height=0.12\linewidth]{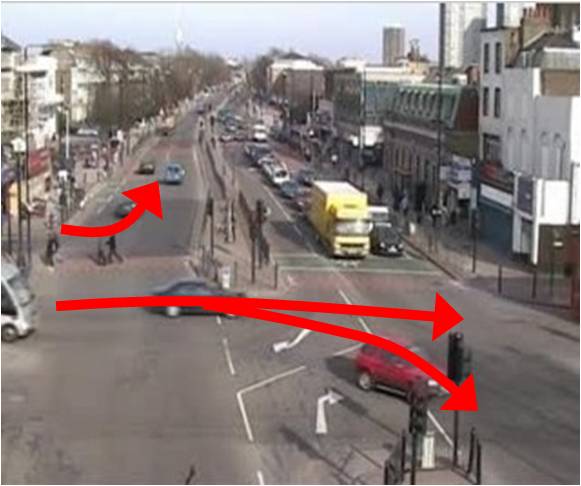}\label{fig:groundtruth_acitivity_l}}
  \hspace{0.1mm}
  \subfloat[VB]{\includegraphics[width=0.185\linewidth,height=0.12\linewidth]{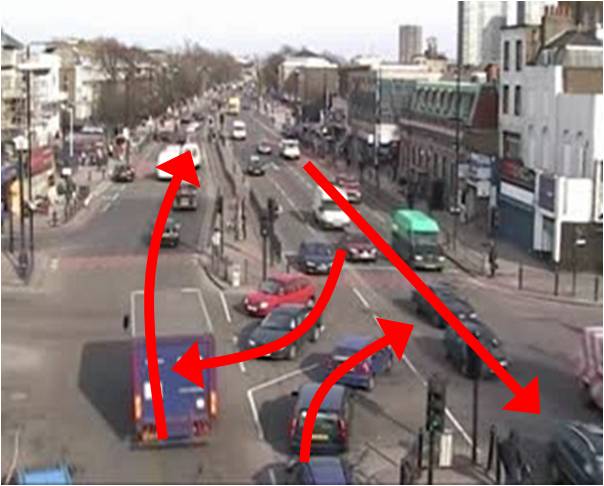}\label{fig:groundtruth_acitivity_m}}
  \caption{Example frames of the ground-truth activities in different videos. First and second rows: videos in our dataset; Third row: video of QMUK Junction dataset \cite{36}. HD: Horizontal pass and down turn; HP: Horizontal pass; BT: Both turn; VP: Vertical pass; VL: Vertical pass and left turn; HU: Horizontal pass and up turn; VR: Vertical pass and right turn; VB: Vertical pass and both turn.}\label{fig:groundtruth_acitivity}
\end{figure}

We perform experiments on two crowd videos in our dataset, as the first and second rows in Fig.~\ref{fig:groundtruth_acitivity}. 400 video clips are selected from each video with each clip including 20 frames. Four crowd activities are defined for each video and the example frames for the crowd activities are shown in Fig.~\ref{fig:groundtruth_acitivity}. Note that these videos are challenging in that: (1) the crowd density in the scene varies frequently including both high density as Fig.~\ref{fig:groundtruth_acitivity_d} and low density clips as Fig.~\ref{fig:groundtruth_acitivity_c}; (2) The motion patterns are varying for different activities, making it difficult to construct meaningful and stable semantic regions; (3) There are large numbers of irregular motions that disturb the normal motion patterns (e.g., people running the red lights or bicycle following irregular paths); (4) The number of clips in the dataset is small, which increases the difficulty of constructing reliable semantic regions. Moreover, in order to further demonstrate the effectiveness of our approach, we also perform experiments on a public QMUK Junction dataset \cite{36} where five crowd activities are defined, as shown in the third row of Fig.~\ref{fig:groundtruth_acitivity}.

\begin{figure}
 \centering
  \subfloat[\scriptsize{Original}]{\includegraphics[width=0.18\linewidth]{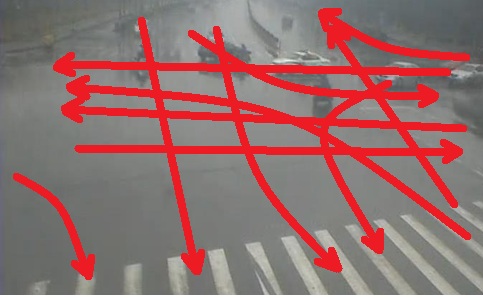}\label{fig:semantic_results_a}}
  \hspace{1mm}
  \subfloat[\scriptsize{Our}]{\includegraphics[width=0.18\linewidth]{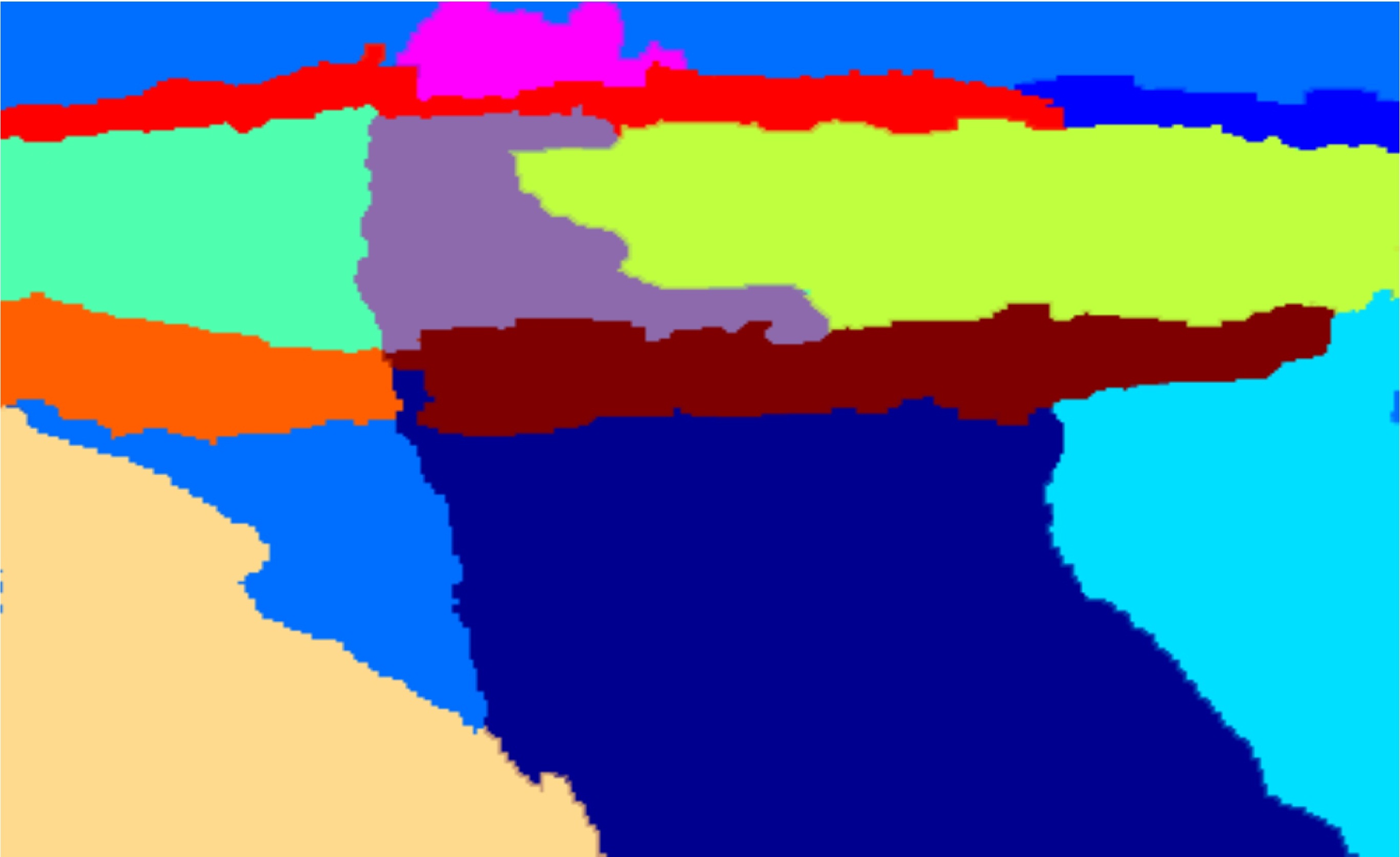}\label{fig:semantic_results_b}}
  \hspace{1mm}
  \subfloat[\scriptsize{Direct}]{\includegraphics[width=0.18\linewidth]{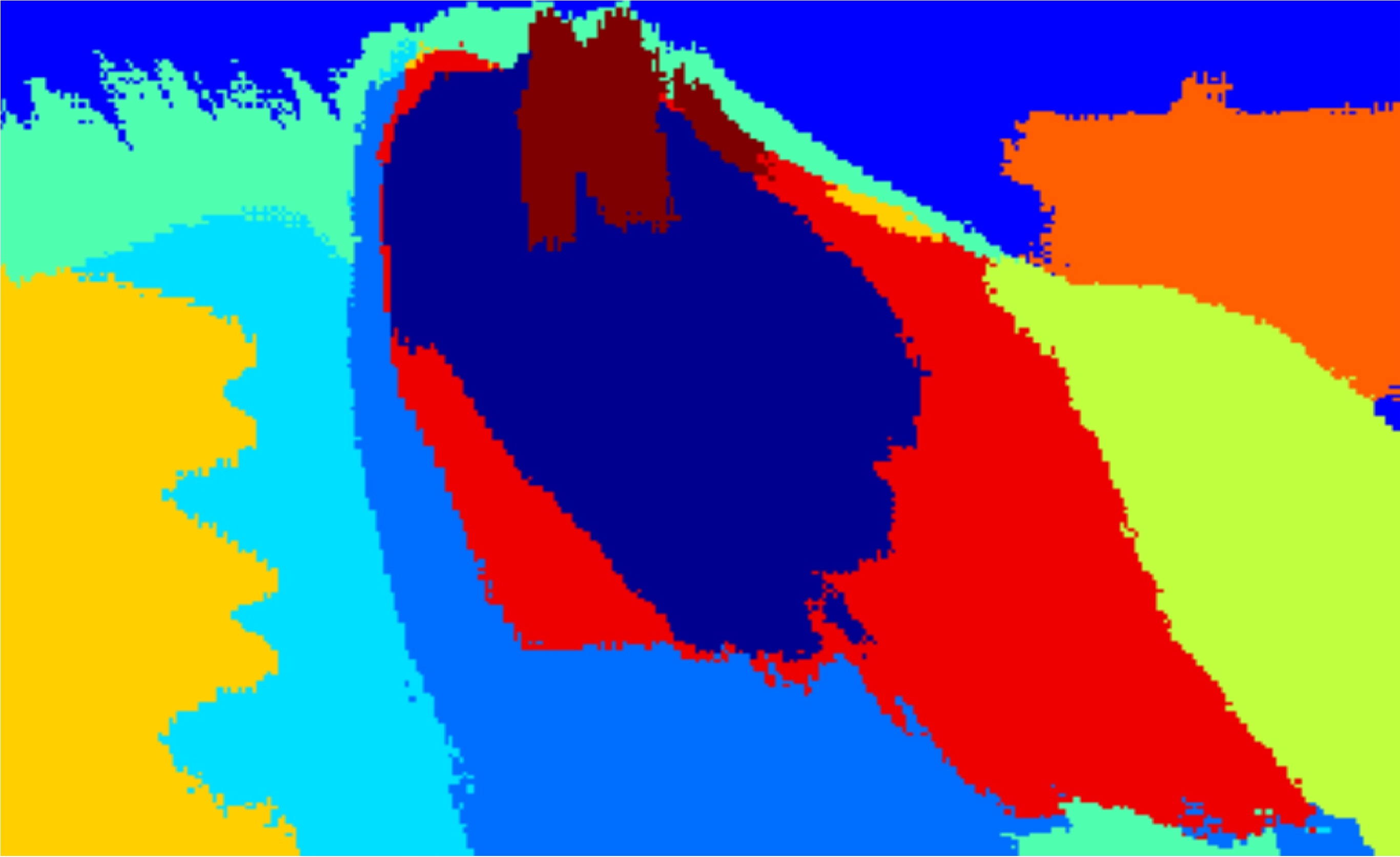}\label{fig:semantic_results_c}}
  \hspace{1mm}
  \subfloat[\scriptsize{\cite{2}}]{\includegraphics[width=0.18\linewidth,height=0.97cm]{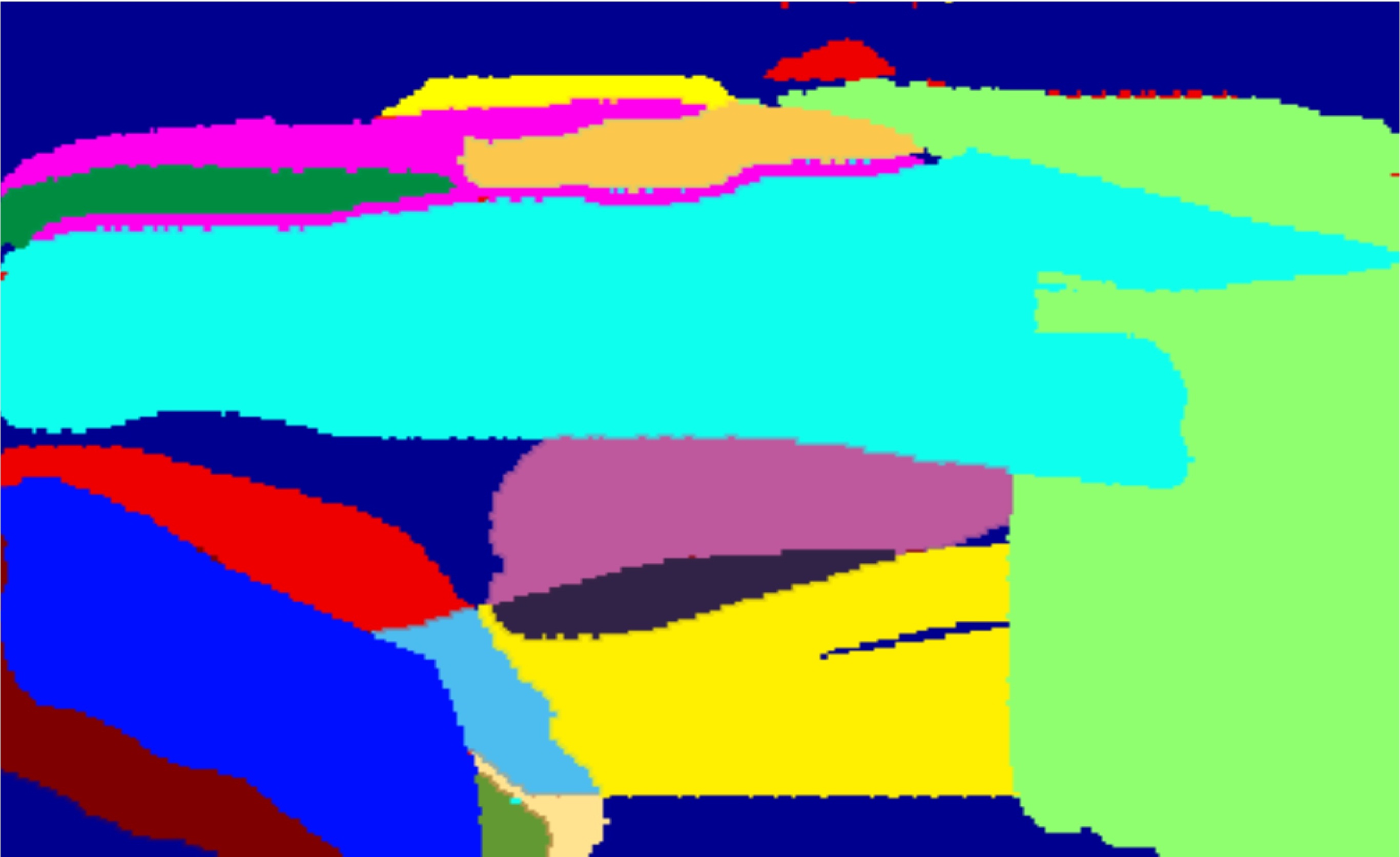}\label{fig:semantic_results_d}}
  \hspace{1mm}
  \subfloat[\scriptsize{\cite{22}}]{\includegraphics[width=0.18\linewidth]{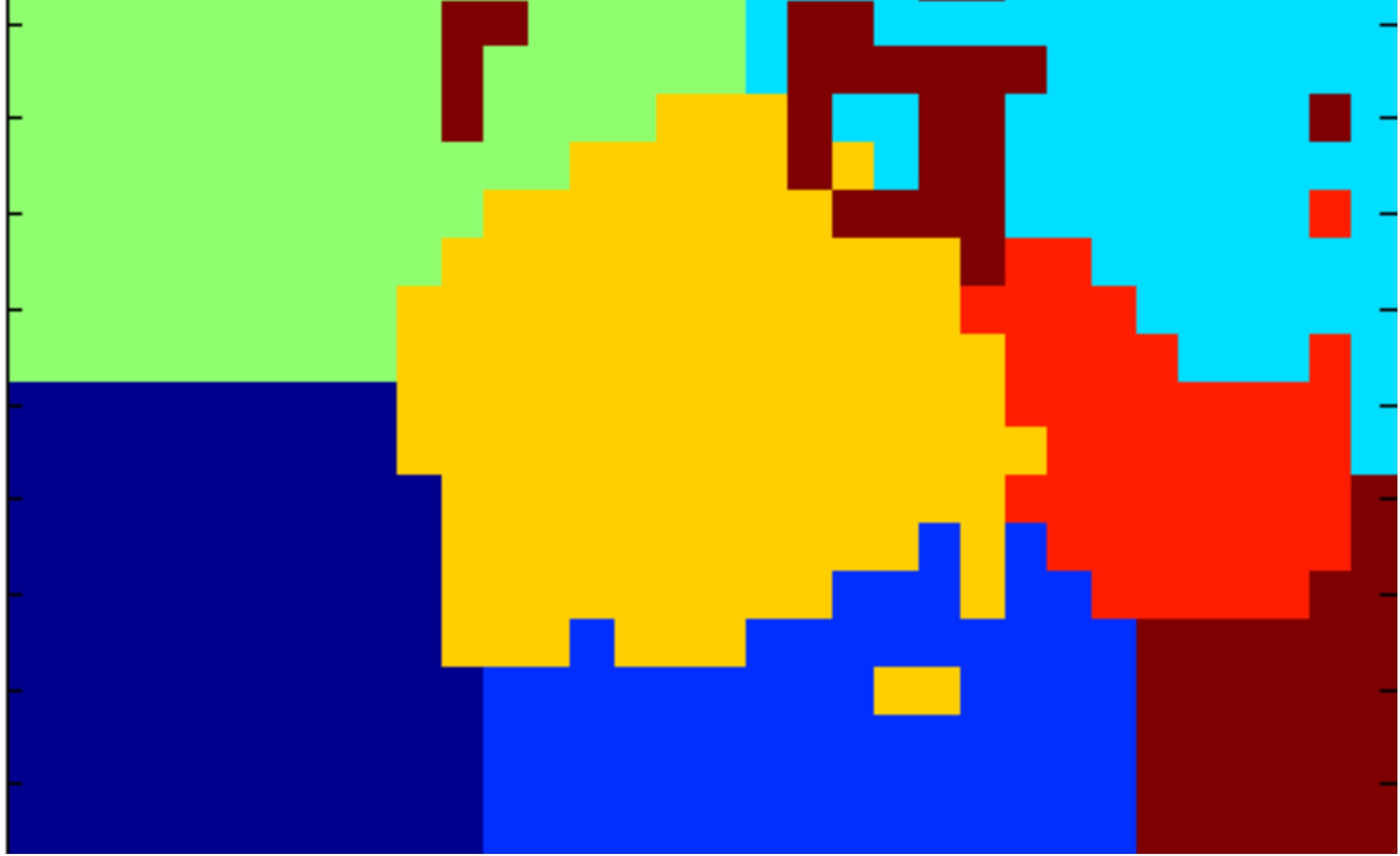}\label{fig:semantic_results_e}}\\
  \subfloat[\scriptsize{Original}]{\includegraphics[width=0.18\linewidth,height=0.97cm]{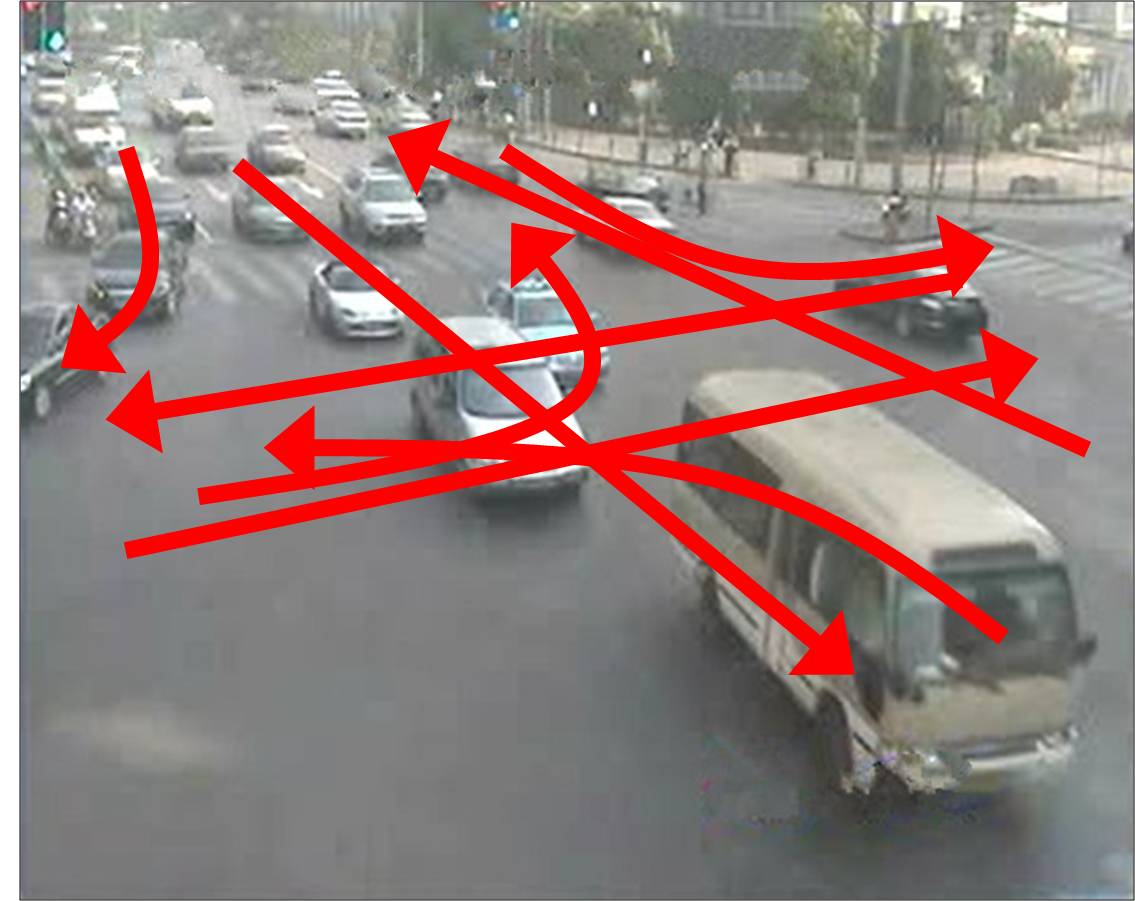}\label{fig:semantic_results_f}}
  \hspace{1mm}
  \subfloat[\scriptsize{Our}]{\includegraphics[width=0.18\linewidth,height=0.97cm]{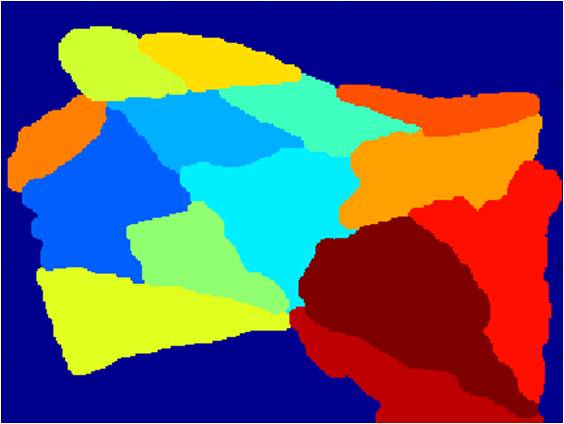}\label{fig:semantic_results_g}}
  \hspace{1mm}
  \subfloat[\scriptsize{Direct}]{\includegraphics[width=0.18\linewidth,height=0.97cm]{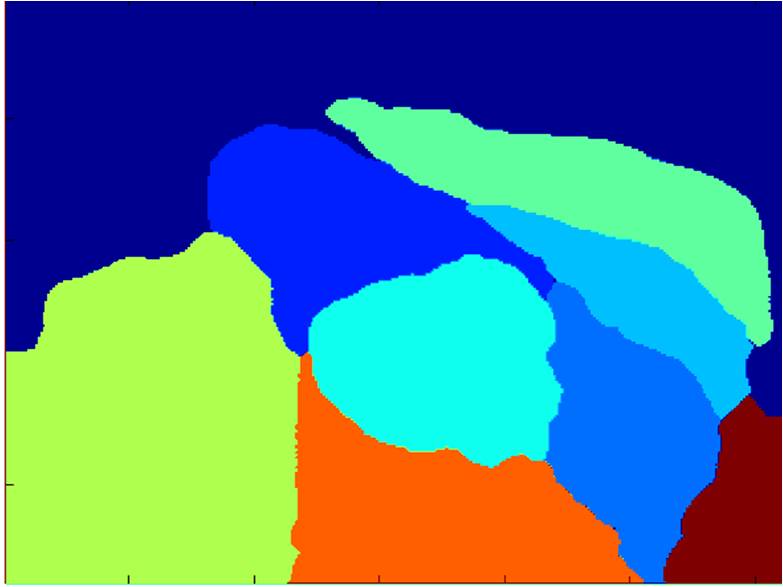}\label{fig:semantic_results_h}}
  \hspace{1mm}
  \subfloat[\scriptsize{\cite{2}}]{\includegraphics[width=0.18\linewidth,height=0.97cm]{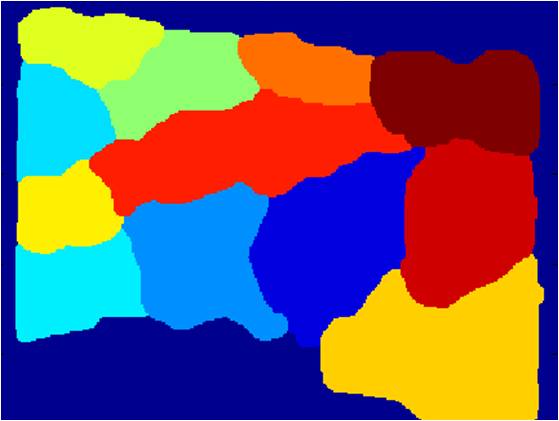}\label{fig:semantic_results_i}}
  \hspace{1mm}
  \subfloat[\scriptsize{\cite{22}}]{\includegraphics[width=0.18\linewidth,height=0.97cm]{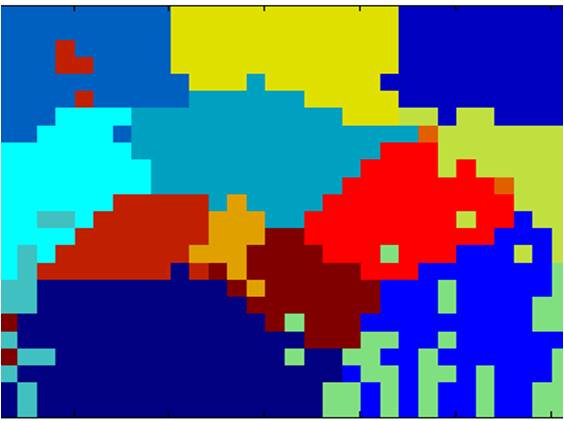}\label{fig:semantic_results_j}}\\
  \subfloat[\scriptsize{Original}]{\includegraphics[width=0.18\linewidth,height=0.97cm]{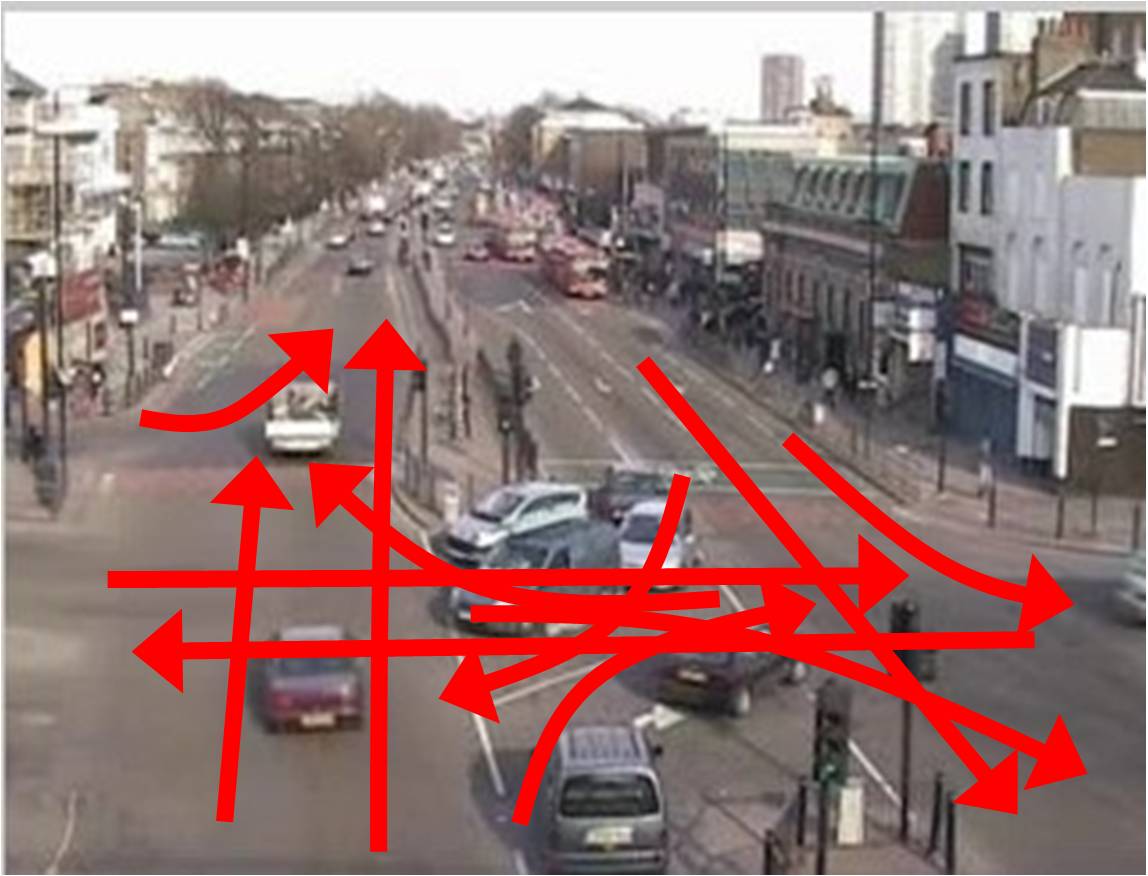}\label{fig:semantic_results_k}}
  \hspace{1mm}
  \subfloat[\scriptsize{Our}]{\includegraphics[width=0.18\linewidth,height=0.97cm]{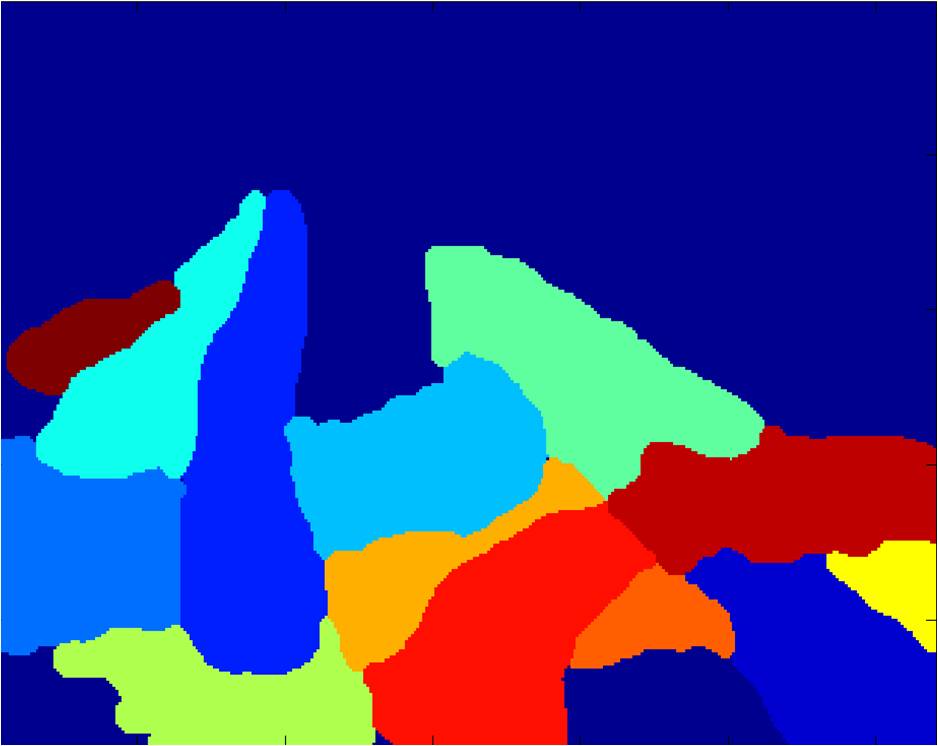}\label{fig:semantic_results_l}}
  \hspace{1mm}
  \subfloat[\scriptsize{Direct}]{\includegraphics[width=0.18\linewidth,height=0.97cm]{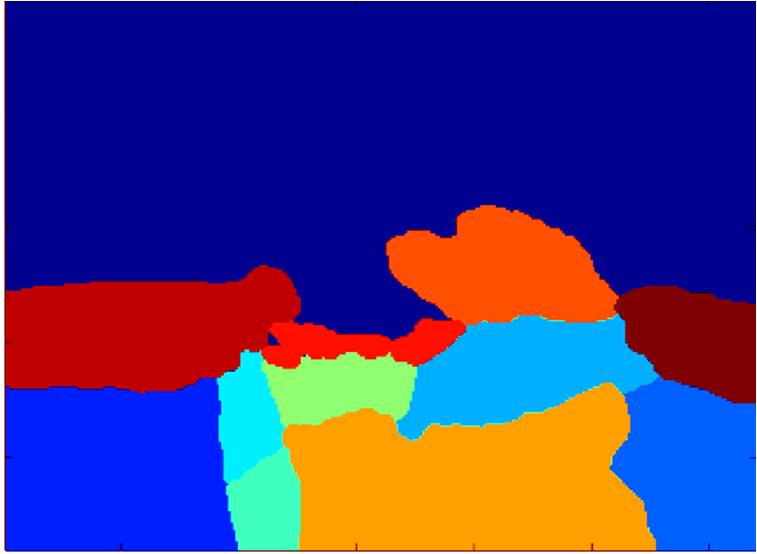}\label{fig:semantic_results_m}}
  \hspace{1mm}
  \subfloat[\scriptsize{\cite{2}}]{\includegraphics[width=0.18\linewidth,height=0.97cm]{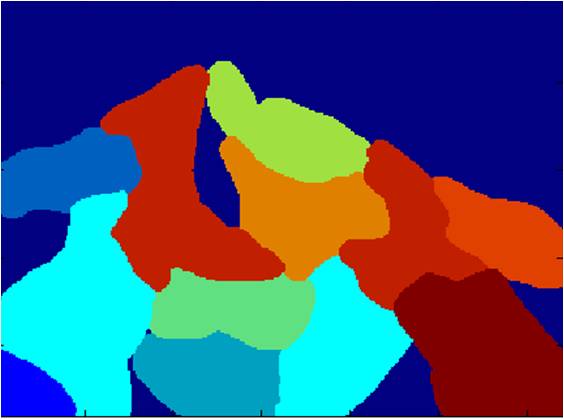}\label{fig:semantic_results_n}}
  \hspace{1mm}
  \subfloat[\scriptsize{\cite{22}}]{\includegraphics[width=0.18\linewidth,height=0.97cm]{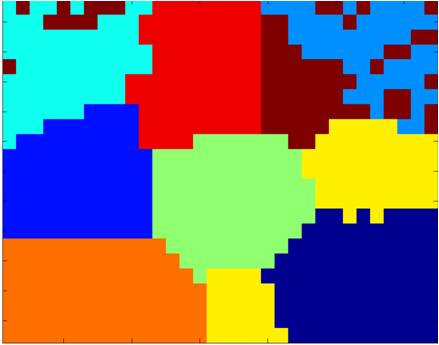}\label{fig:semantic_results_o}}
 \caption{Constructed semantic regions of different methods for the videos in Fig.~\ref{fig:groundtruth_acitivity}. The caption ``\cite{2}'' denotes the method ``\cite{2}+Two step''. (Best viewed in color)}\label{fig:semantic_results}
\end{figure}

{\bf Accuracy on semantic region construction.} For each video in Fig.~\ref{fig:groundtruth_acitivity}, we randomly select 200 video clips and use them to construct the corresponding semantic regions. Fig.~\ref{fig:semantic_results} compares the results of four methods: (1) Our approach (``Our''), (2) Directly cluster regions based on the particles' TEF vectors (``Direct'', note that our approach differs from this method by clustering over the cluster label vectors), (3) Use \cite{2} to achieve coherent motion regions and then apply our two-step clustering scheme to construct semantic regions (``\cite{2}+Two-Step'', we show the results of \cite{2} because in our experiments, \cite{2} has the best semantic region construction results among the compared methods in Table~\ref{table:table_CNE}), (4) The activity-based scene segmentation method in \cite{22} (``\cite{22}''). We also show original scene images and plot all major activity flows to ease the comparison (``original scene'').

Fig.~\ref{fig:semantic_results} shows that the methods utilizing ``coherent motion cluster label'' information (``our'' and ``\cite{2}+two-step'') create more meaningful semantic regions than the other methods, e.g., successfully identifying the horizontal motion regions in the middle of the scene in Fig.~\ref{fig:semantic_results_b}. This shows that our cluster label features can effectively strengthen the correlation among particles to facilitate semantic region construction. Furthermore, comparing our approach with the ``\cite{2}+Two-Step'' method, it is obvious that the semantic regions by our approach are more accurate (e.g., more precise semantic region boundaries and more meaningful segmentations in the scene). This further shows that more precise coherent motion detection results can result in more accurate semantic region results.

{\bf Performances on recognizing pre-defined activities.} In order to recognize the pre-defined activities in Fig.~\ref{fig:groundtruth_acitivity}, for each video, we randomly select 200 video clips and construct semantic regions by the methods in Fig.~\ref{fig:semantic_results}. After that, we derive features from the TEF and train SVM classifiers by the method in Section~\ref{section:pre_defined_recognition}. Finally, we perform recognition on the other 200 video clips in the same video. Besides, we also include the results of two additional methods: (1) a state-of-the-art dense-trajectory-based recognition method \cite{20} (``Dense-Traj''); (2) the method which uses our semantic regions but uses the input motion field (i.e., the optical flows) to derive the motion features in each semantic region (``Our+OF''). From the recognition accuracy shown in Table~\ref{table:table_activity}, we observe that:
\begin{enumerate}
 \item Methods using more meaningful semantic regions (i.e., ``our'', ``our+OF'', and ``\cite{2}+Two step'') achieve better results than other methods. This shows that suitable semantic region construction can greatly facilitate activity recognition.
 \item Approaches using TEF (``Our'') achieve better results than those using the input motion field (``Our+OF''). This demonstrates that compared with the input motion filed, our TEF can effectively improve the effectiveness in representing the semantic regions' motion patterns.
 \item The dense-trajectory method \cite{20} which extracts global features does not achieve satisfying results. This is because the global features still have limitations in differentiating the subtle differences among activities. This further implies the usefulness of semantic region decomposition in analyzing crowd scenes.
\end{enumerate}

\begin{table}
\caption{\label{comparison2}Recognition accuracy of different methods}
  \centering
  \newcommand{\tabincell}[2]{\begin{tabular}{@{}#1@{}}#2\end{tabular}}
  \begin{tabular}{|c|c|c|c|c|c|c|c|}
    \hline
      &\tabincell{c}{\bf{Our}\\ \bf{(\%)}}&\tabincell{c}{Our+\\OF (\%)}&\tabincell{c}{Direct\\ (\%)}&\tabincell{c}{\cite{2}+Two-\\Step (\%)}&\tabincell{c}{\cite{22}\\ (\%)}&\tabincell{c}{\cite{20}\\(\%)}\\
    \hline
      \tabincell{c}{Fig.~\ref{fig:groundtruth_acitivity_a} video} &\bf{92.2}&87.75&77.0&89.5&79.2&67.0\\
      \tabincell{c}{Fig.~\ref{fig:groundtruth_acitivity_e} video } &\bf{90.69}&83.83&73.53&81.76&72.35&69.80\\
      \tabincell{c}{Fig.~\ref{fig:groundtruth_acitivity_i} video} &\bf{93.58}&91.03&83.42&88.32&84.83&82.69\\
    \hline
  \end{tabular}\label{table:table_activity}
\end{table}

\subsection{Results for Recurrent Activity Mining}

In this experiment, we use the same videos as in Fig.~\ref{fig:groundtruth_acitivity} for mining recurrent activities. For each video, we sample one frame per second, then calculate coherent motions for the sampled frames, and finally apply our cluster-and-merge process to achieve recurrent activity patterns. Note that the target for recurrent activity mining is to automatically discover recurrent activities from an input video without pre-defining activity types or pre-labeling training data. And ideally, good activity mining approaches should achieve similar activity patterns as the human-observed activity types in Fig.~\ref{fig:groundtruth_acitivity}.

{\bf Performances on frame-level clustering.}
For each video, we apply our frame-level clustering step to cluster the sampled frames into four recurrent activity groups. Our clustering results are compared with two methods:
(1) \emph{Direct clustering.} Directly clustering based on the TEF difference between two frames (i.e., use the summation of absolute thermal energy differences between the co-located particles in two TEFs as the inter-frame similarity).
(2) \emph{Pre-clustering.} Using the matched-coherent-motion similarities $S_{FM}\left(t,t-\tau\right)$ in Eq.~\ref{equation:eq9} as the inter-frame similarity for clustering.

\begin{figure}
 \centering
  \subfloat[\scriptsize{Our}]{\includegraphics[width=0.25\linewidth,height=2.2cm]{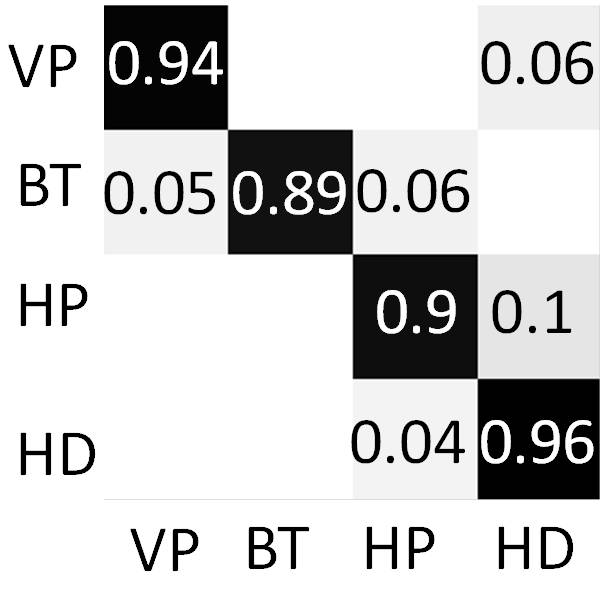}}
  \hspace{6mm}
  \subfloat[\scriptsize{Direct Clustering}]{\includegraphics[width=0.25\linewidth,height=2.2cm]{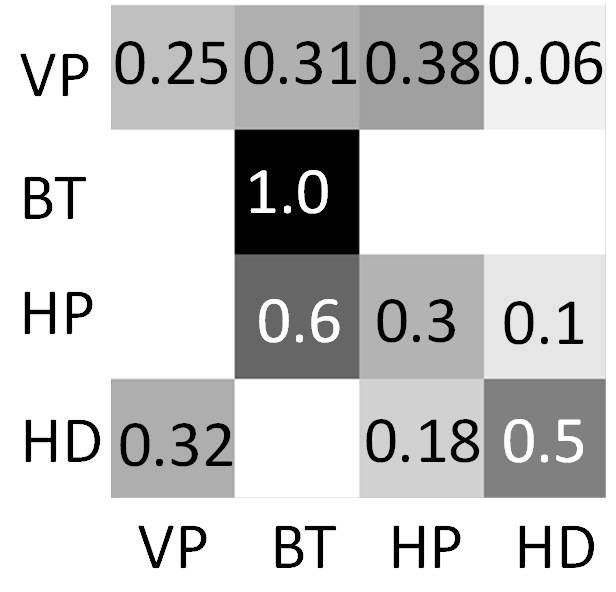}}
  \hspace{6mm}
  \subfloat[\scriptsize{Pre-clustering}]{\includegraphics[width=0.25\linewidth,height=2.2cm]{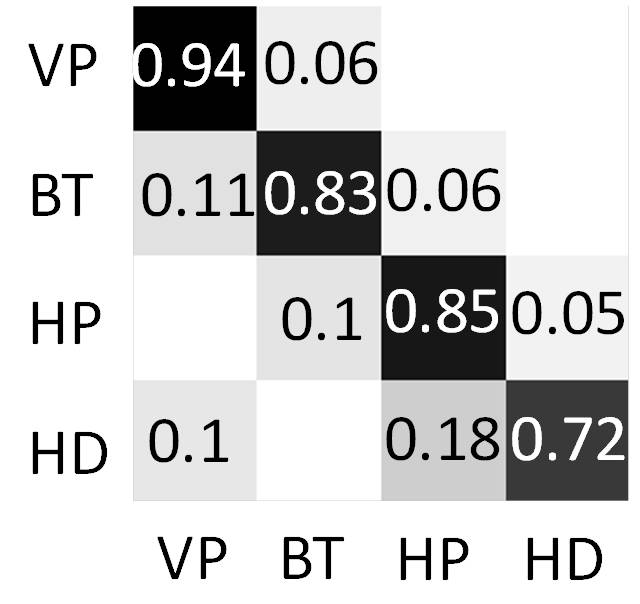}}\\
  \subfloat[\scriptsize{Our}]{\includegraphics[width=0.25\linewidth,height=2.2cm]{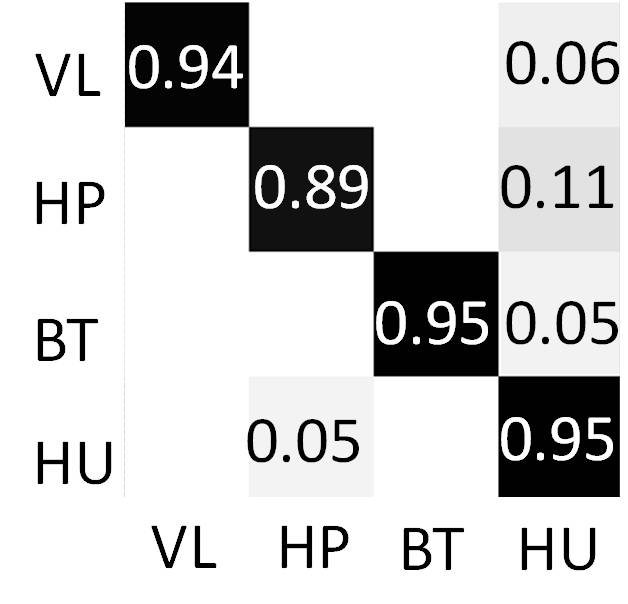}}
  \hspace{6mm}
  \subfloat[\scriptsize{Direct Clustering}]{\includegraphics[width=0.25\linewidth,height=2.2cm]{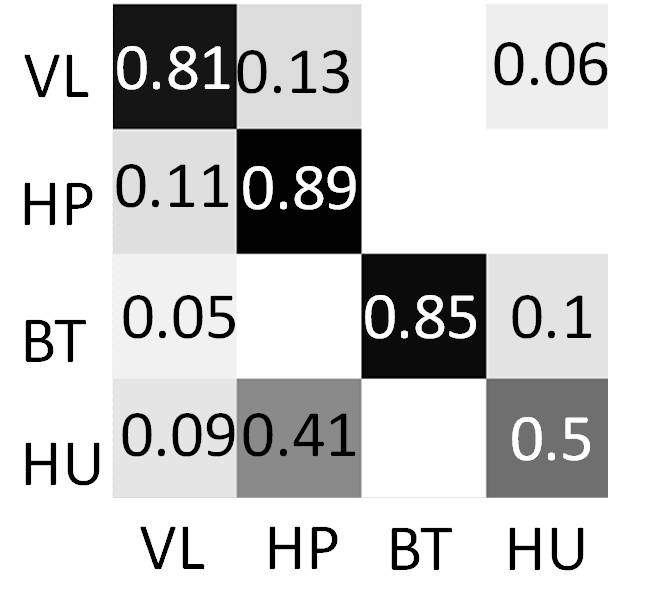}}
  \hspace{6mm}
  \subfloat[\scriptsize{Pre-clustering}]{\includegraphics[width=0.25\linewidth,height=2.2cm]{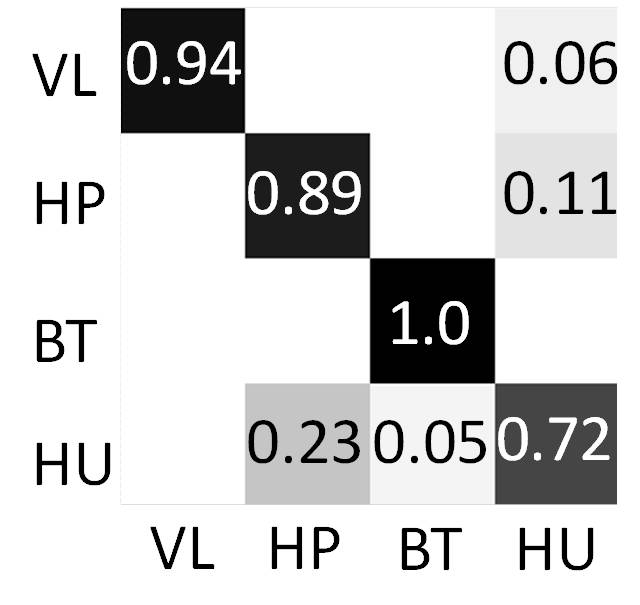}}\\
  \subfloat[\scriptsize{Our}]{\includegraphics[width=0.29\linewidth,height=2.5cm]{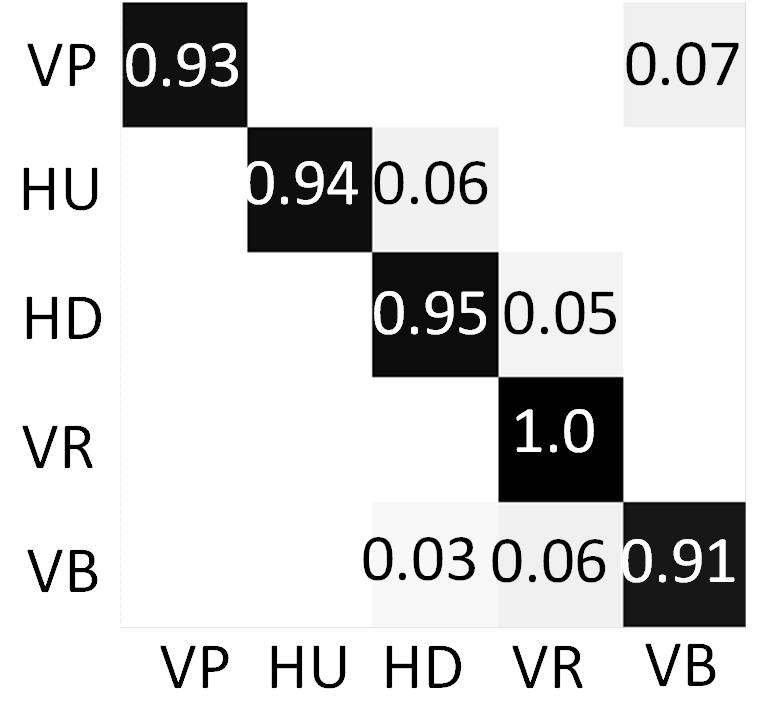}}
  \hspace{2mm}
  \subfloat[\scriptsize{Direct Clustering}]{\includegraphics[width=0.29\linewidth,height=2.5cm]{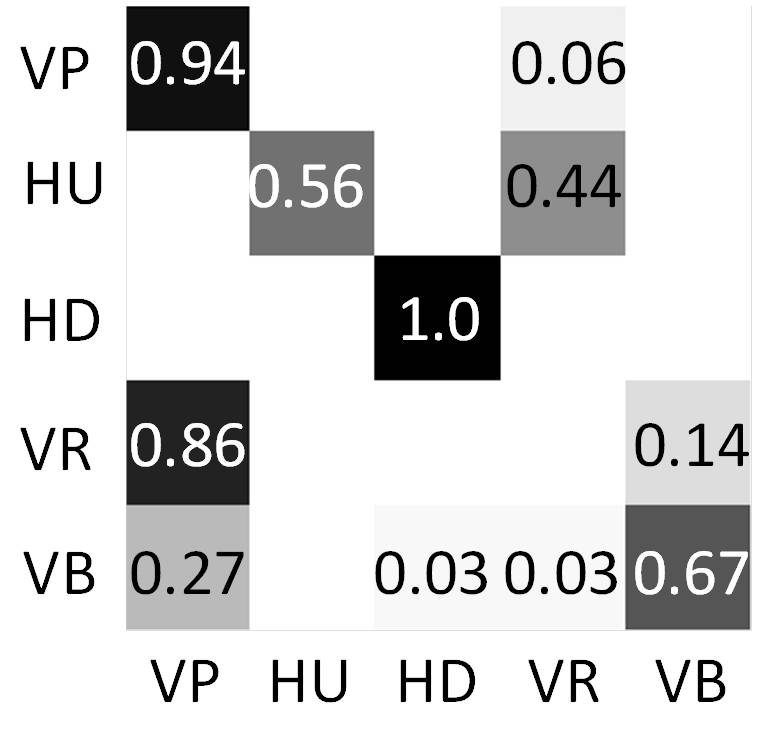}}
  \hspace{2mm}
  \subfloat[\scriptsize{Pre-clustering}]{\includegraphics[width=0.29\linewidth,height=2.5cm]{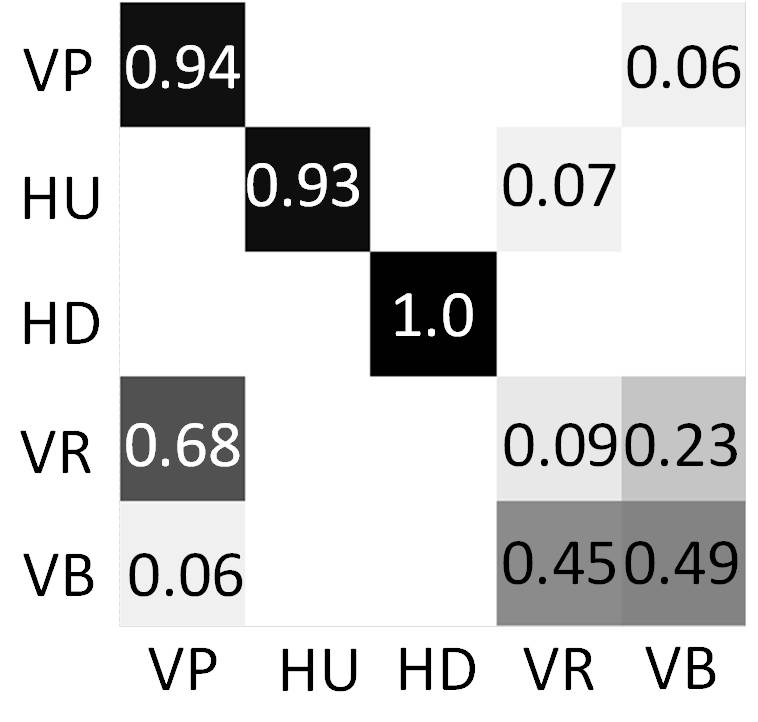}}
  \caption{The confusion matrix of clustering results. (The clustering results in the 1st, 2nd, and 3rd rows correspond to the videos in the 1st, 2nd, and 3rd rows in Fig.~\ref{fig:groundtruth_acitivity}, respectively)}\label{fig:confusion_matrix}
\end{figure}

Fig.~\ref{fig:confusion_matrix} compares the clustering confusion matrixes of different methods. From Fig.~\ref{fig:confusion_matrix}, we can see that since frames of the same recurrent activity may contain different parts of a complete activity flow (e.g., Fig.~\ref{fig:coherent_merging}), their TEFs may have large differences. Therefore, directly using TEF difference for clustering (i.e., direct TEF clustering) cannot achieve satisfying results. Comparatively, by including coherent motions to evaluate inter-frame similarities (i.e., ``pre-clustering'' and ``our''), the clustering accuracy can be improved. However, the pre-clustering method still have limitations in differentiating similar recurrent activities, e.g., HP and HU in Figs~\ref{fig:groundtruth_acitivity_g} and~\ref{fig:groundtruth_acitivity_h}. Comparatively, by introducing the importance cost of semantic regions to measure the effects of unmatched coherent motions, our frame-level clustering approach can have stronger capability in differentiating similar recurrent activity patterns.

\begin{figure*}
 \centering
 \subfloat[\scriptsize{Recur. act. mining results for video of Fig.~\ref{fig:groundtruth_acitivity_a}}]{\includegraphics[width=0.3\linewidth,height=0.169\linewidth]{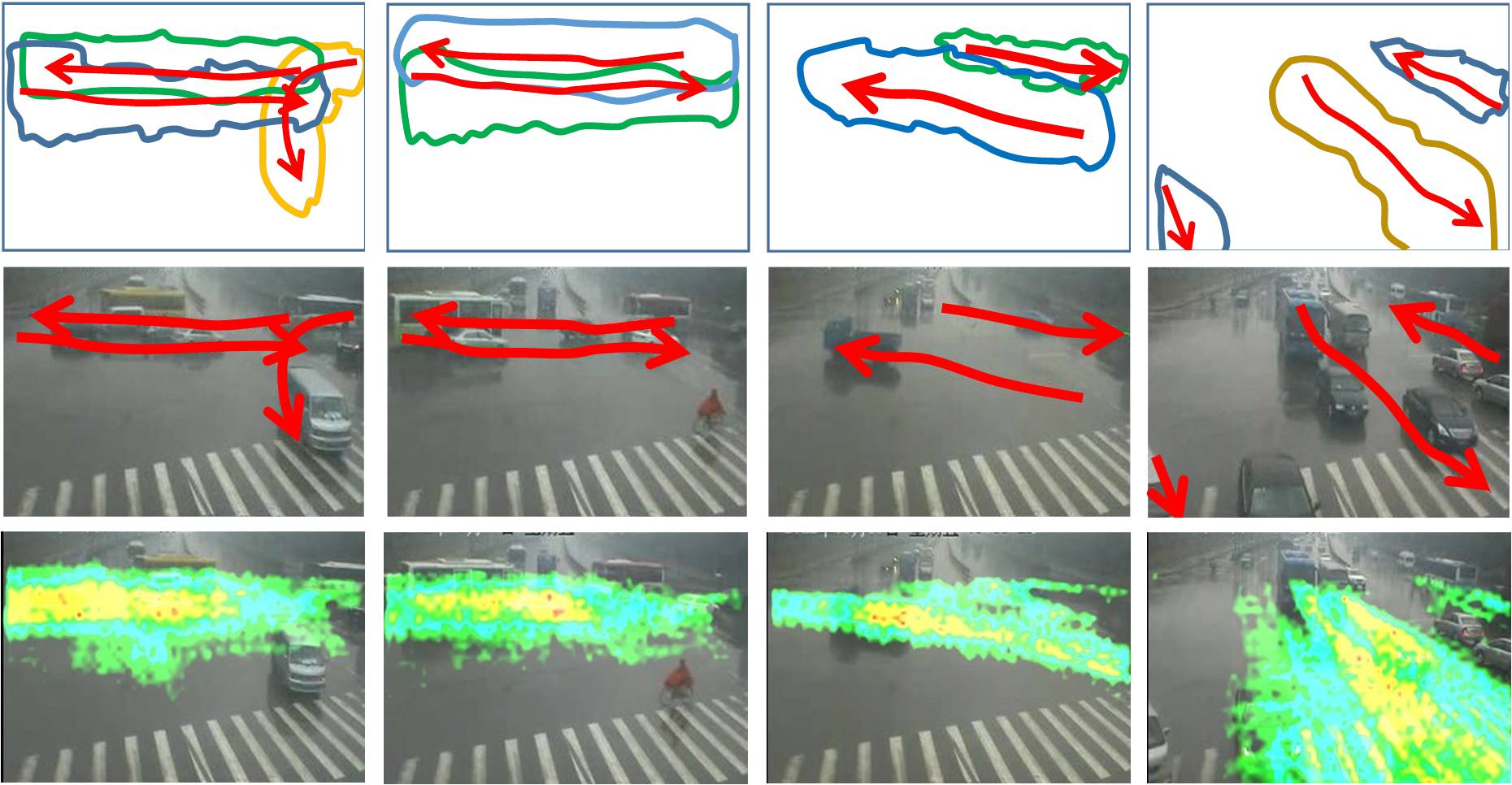}\label{fig:activity_mining_results_a}}
 \hspace{3pt}
 \subfloat[\scriptsize{Recur. act. mining results for video of Fig.~\ref{fig:groundtruth_acitivity_e}}]{\includegraphics[width=0.3\linewidth,height=0.169\linewidth]{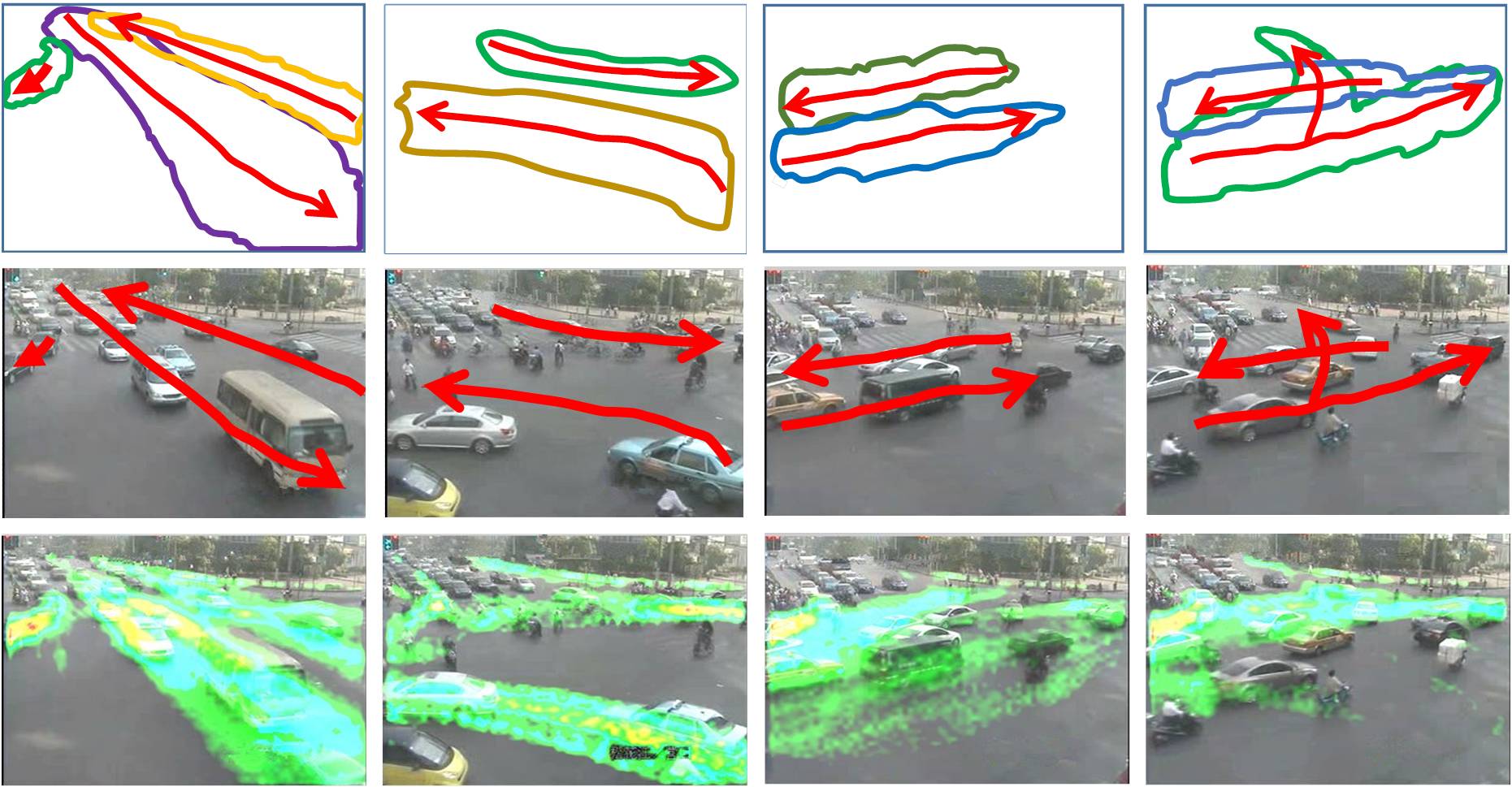}\label{fig:activity_mining_results_b}}
 \hspace{3pt}
 \subfloat[\scriptsize{Recur. act. mining results for video of Fig.~\ref{fig:groundtruth_acitivity_i}}]{\includegraphics[width=0.37\linewidth,height=0.169\linewidth]{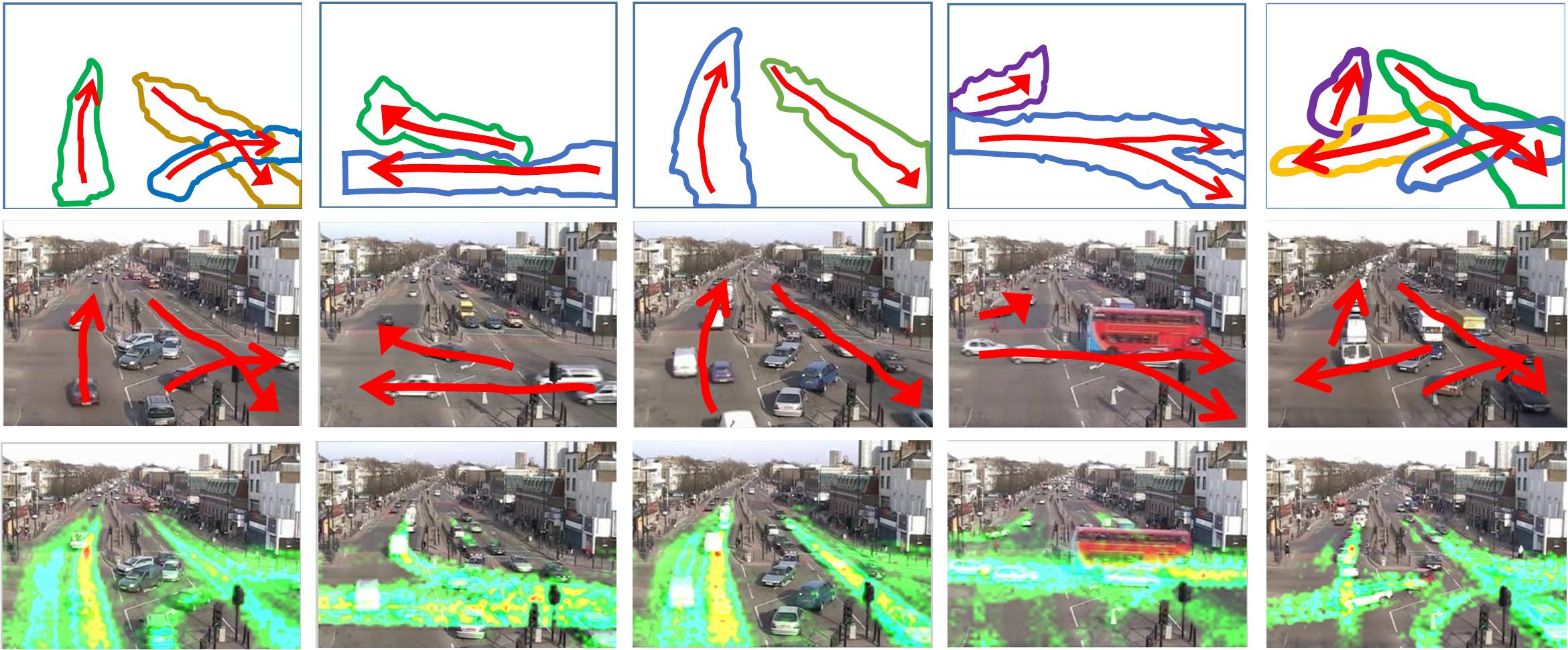}\label{fig:activity_mining_results_c}}
 \caption{Recurrent activity mining results. First rows in (a), (b), and (c): the merged motion pattern regions and the extracted flow curves by our approach; Second rows in (a), (b), and (c): the extracted flow curves of our approach displayed over the video frame; Third rows in (a), (b), and (c): the recurrent activities extracted by \cite{32}.}\label{fig:activity_mining_results}
\end{figure*}

{\bf Performances on coherent motion merging and flow curve extraction.}
Fig.~\ref{fig:activity_mining_results} shows the results of our coherent motion merging and flow curve extraction steps. Besides, we also compare our approach with a state-of-the-art activity mining method which utilizes a Probabilistic Latent Sequential Motif (PLSM) model to discover recurrent activities \cite{32}, which are shown as the last rows in Figs~\ref{fig:activity_mining_results_a},~\ref{fig:activity_mining_results_b}, and~\ref{fig:activity_mining_results_c}. From Fig.~\ref{fig:activity_mining_results}, we can have the following observations:
\begin{enumerate}
 \item The recurrent activities mined by our approach is similar to the human-observed activity types in Fig.~\ref{fig:groundtruth_acitivity}. This demonstrates that our proposed cluster-and-merge process can effectively discover desired activity types from an input video.
 \item Note that although the clustering result in our frame-level clustering step is not 100 percentage accurate (as in Fig.~\ref{fig:confusion_matrix}), the extracted flow curves are less affected by the wrongly clustered frames because: (i) The noisy or isolated thermal energy vectors from the wrongly clustered frames will be filtered by the threshold $\theta_{mf}$ in Eq.~\ref{equation:eq13}. (ii) The flow curve extraction process will further reduce the effects of wrong frames by dividing sub-regions to derive flow curves, as in Fig.~\ref{fig:curve_extraction_a}.
 \item Comparing our approach with the PLSM-based method \cite{32}, we can see that: (i) By introducing coherent regions to measure inter-frame similarities and derive motion pattern regions, our approach can achieve cleaner activity flows which are more coherent with the human-observed activity types in Fig.~\ref{fig:groundtruth_acitivity}. Comparatively, results of the PLSM-based method still include noisy motion patterns, e.g., the last column in Fig.~\ref{fig:activity_mining_results_b}. (ii) Our approach can precisely differentiate motion flows inside a recurrent activity. However, the PLSM-based method has limitations in differentiate motion flows when they are located close to each other, e.g., the second column in Fig.~\ref{fig:activity_mining_results_a}. (iii) The differences between similar recurrent activities are clearly differentiated and visualized by our approach, while they are less obvious in the results of the PLSM-based method, e.g., the third and fourth columns in Fig.~\ref{fig:activity_mining_results_b}.
\end{enumerate}

\section{Conclusion\label{section:conclusion}}

In this paper, we study the problem of coherent motion detection, semantic region construction, and recurrent activity mining in crowd scenes. A thermal-diffusion-based algorithm together with a two-step clustering scheme are introduced, which can achieve more meaningful coherent motion and semantic region results. Based on the extracted coherent motions and semantic regions, a cluster-and-merge process is further proposed which can effectively discover desirable activity patterns from a crowd video. Experiments on various videos show that our approach achieves the state-of-the-art performance.

\bibliographystyle{IEEEtran}
\bibliography{egbib}

\end{document}